\documentclass{article}
\usepackage[utf8]{inputenc}
\usepackage{graphicx}
\usepackage{amsmath}
\usepackage{amsthm}
\usepackage{amsfonts}
\usepackage{xcolor}
 \usepackage{relsize}
\usepackage[font=small]{caption}
\usepackage{algorithm}
\usepackage{subcaption}
\newtheorem{theorem}{Theorem}
\title{Stochastic Operator Network: A Stochastic Maximum Principle Based Approach to Operator Learning}
\author{Ryan Bausback, Jingqiao Tang, Lu Lu, Feng Bao, Toan Huynh}

\begin{document}

\maketitle

\section{Abstract}
We develop a novel framework for uncertainty quantification in operator learning, the Stochastic Operator Network (SON). SON combines the stochastic optimal control concepts of the Stochastic Neural Network (SNN) with the DeepONet. By formulating the branch net as an SDE and backpropagating through the adjoint BSDE, we replace the gradient of the loss function with the gradient of the Hamiltonian from Stohastic Maximum Principle in the SGD update. This allows SON to learn the uncertainty present in operators through its diffusion parameters. We then demonstrate the effectiveness of SON when replicating several noisy operators in 2D and 3D.

\section{Introduction}

{\color{black} Over the past five years, operator learning has emerged as a promising alternative to traditional numerical solvers for a wide range of differential equations~\cite{Bhattacharya2021, Chen1995, Guo2024, Kovachki2024,  Lee2024, Li2020, Lu2022,Rahman2022,  Zhang2022}. Unlike classical machine learning models that approximate functions at discrete spatial or temporal points, neural operators take input as an entire function and produce a corresponding output function~\cite{Lu2019}. The two most influential architectures, Deep Operator Network (DeepONet) and Fourier Neural Operator (FNO), have driven the evolution of deep operator learning. DeepONet, which is based on the Universal Approximation Theorem for Operators~\cite{Chen1995}, comprises two parallel subnetworks: the ``branch'' net, which learns coefficients, and the ``trunk'' net, which learns a data‐driven basis for the output function. FNO, in contrast, encodes and decodes using the Fourier basis (via FFT and inverse FFT) within successive Fourier layers, relying on a single feed‐forward backbone and assuming that input and output domains coincide~\cite{Li2020}. The authors in~\cite{Lu2022} later showed that FNO was simply a special parameterization of DeepONet under the Fourier basis, which means that all variants of the FNO benefits from the same universal approximation feature as the DeepONet~\cite{Lu2022}. Crucially, neither DeepONet nor FNO relies on a fixed mesh discretization, a key limitation of many classical PDE solvers in solving large scale problems, which paves the way for rapid, mesh‐free inference across a broad class of parametrized PDEs.}

{\color{black} While DeepONet has been applied to certain stochastic problems, such as stochastic ODEs and stochastic PDEs in~\cite{Lu2021}, the randomness appears only in the input function, not as intrinsic stochasticity in the operator itself. The network still learns a deterministic map from noisy inputs to outputs, rather than an operator that injects noise at each point along a trajectory. Beyond these stochastic input applications, several advanced operator learning frameworks have emerged. For example, GANO (Generative Adversarial Neural Operator) uses an FNO backbone to train a generator–discriminator pair for richer output sampling, and PCAnet replaces the Fourier basis with a data‐driven PCA basis to reduce dimensionality~\cite{Bhattacharya2021, Rahman2022}. A handful of architectures explicitly target uncertainty quantification: BelNet places a Bayesian prior on DeepONet’s weights to quantify epistemic uncertainty~\cite{Zhang2022}, while IB‐UQ replaces the branch network with an encoder–decoder that learns a latent representation of input noise to produce predictive distributions~\cite{Guo2024}. However, most operator‐learning work still focuses on deterministic accuracy, and even UQ‐oriented variants have not yet tackled SDE‐style operators with pointwise noise along trajectories.}

{\color{black} In this work, we propose a novel training strategy for stochastic operators by merging probabilistic learning, namely, Stochastic Neural Networks (SNNs), with the DeepONet framework. The SNN is an extension of the so-called ``neural ODE" network architecture
where the evolution of hidden layers in the DNN is formulated as a discretized
ordinary differential equation (ODE) system~\cite{Chen2018, Dupont2019, Gerstberger1997, Haber2018, Weinan2017}. More specifically, an additive Brownian
motion noise, which characterizes the randomness
caused by the uncertainty of models and noises of data, is added to the ODE system corresponding to the hidden layers to transform the ODE into an SDE~\cite{Jia2019, Kong2020, Liu2019, Liu2020,  Tzen2019}. In the SNN model,  the prediction of the network is represented through the drift parameters, and the stochastic diffusion governs the randomness of network output, which serves to quantify the uncertainty of deep learning. Compared to other probabilistic learning methods, such as the Bayesian neural network (BNN)~\cite{Geneva2019, Geneva2020, Kwon2020, McDermott2019, Savchenko2020, Wu2020, Yang2021, Yao2019}, the SNN approach takes less computational cost to evaluate the diffusion coefficient, while it is still able to characterize sufficient probabilistic
behavior of the neural network by stacking (controlled) diffusion terms together through the multilayer structure~\cite{Archibald2024}. The main challenge in implementing the SNN is to construct an efficient numerical solver for the
backpropagation process, since the standard backpropagation approach used in the Neural ODE is deemed computationally inefficient~\cite{Archibald2024, Bao2022}. We address this by adopting the sample‐wise backward‐SDE solver of~\cite{Archibald2024,Bao2022}, which treats backward samples as “pseudo data” and solves only a small, randomly selected subset of the backward SDE per iteration, thus drastically reducing computational burden.

The main contributions of our paper are as follows. We first construct the general framework of our training strategy, namely the Stochastic Operator Network (SON), which effectively combines the methodology of the DeepONet and the training process of the SNN. The Stochastic Maximum Principle (SMP)~\cite{Chen2018} is adopted to formulate the loss function of the new backpropagation process and transforms the training algorithm of the neural operator into a stochastic optimal control problem. We then consider several numerical experiments to validate our method on a range of stochastic operators. Finally, we also compare the performance our method with the standard DeepONet approach.}

The rest of this paper is organized as follows: We will detail the intricacies of both methods forming the foundation of SON, elaborating on DeepONet in Section \ref{OpLearn}, and SNN in Section \ref{SNN}. In Section \ref{SON}, we introduce the Stochastic Operator Network and discuss the details of how SNN and DeepONet work in tandem to produce SON. Finally in Section \ref{EXP} we prove the effectiveness of this new architecture by applying it to several noisy 2D and 3D operators, with similar outputs to stochastic differential equations. We also compare SON performance with that of the vanilla DeepONet. 

\section{Deep Operator Network} \label{OpLearn}

{\color{black} Deep Operator Network (DeepONet) is among the most well-known operator learning architectures, whose core concept is to provide an estimation to a given mapping between two Banach spaces~\cite{Kovachki2024}. Given an operator $G$ taking an input function $u$ from some input space $V$, DeepONet returns an approximation operator for $G(u)$ over the set of points $y$ in the domain of $G(u)$. Hence, the network takes inputs composed of two parts: u and y, and outputs $G(u)(y)$~\cite{Lu2021}. To enable the training procedure of the approximation network, we discretize the input function 
$u$ by sampling its values at a finite set of $m$ locations $\{x_1, \hdots, x_m\}$ called ``sensors". Our discretization for any $u$ is therefore $\{u(x_0),..., u(x_m)\}$. Although various representations are possible (see, e.g.,~\cite{Lu2021}), we adopt this simplest approach unless stated otherwise. In our experiments, we use the same $m$ sensors for every input, but more flexible schemes where sensor locations vary by sample have also been explored (see, e.g.,~ \cite{Zhang2022}).

The theoretical foundation of operator learning is the following Universal Approximation Theorem for Operators, first discovered by the authors in~\cite{Chen1995}:

\begin{theorem}[\textbf{Universal Approximation Theorem for Operators}~\cite{Chen1995}]
\label{univapproxop}
Suppose $\sigma$ is continuous non-polynomial function, $X$ is Banach space, $K_1 \subset X$, $K_2 \subset \mathbb{R}^d$ are compact sets, $V$ is compact in $C(K_1)$, and $G$ is operator mapping $V$ to $C(K_2)$. Then for $\epsilon > 0$, there are positive integers $n, p and m$,constants $c^k_i$, $\xi^k_{ij}$,$\theta^k_i$ $\in \mathbb{R}$, $w_k \in \mathbb{R}^d$, $x_j \in K_1$, $i = 1, \hdots, n$,
$k = 1,\hdots, p$ and $j = 1,\hdots, m$, such that
$$\left|G(u)(y) - \sum^p_{k=1} \underbrace{\sum^n_{i=1}c^k_i \sigma\left(\sum^m_{j=1} \xi_{ij}^ku(x_j)+\theta_i^k\right)}_{\text{Branch}}\underbrace{\sigma(w_ky+ \zeta_k)}_{\text{Trunk}}\right| < \epsilon, $$
holds for all $u \in V$ and $y \in K_2$. Here, $C(K)$ is the Banach space of all continuous
functions defined on $K$ with norm $\vert\vert f \vert\vert_{C(K)} = \max\limits_{x \in K} \vert f(x) \vert.$
\end{theorem}

}
{\color{black}
While this result ensures that a shallow branch–trunk pair can in principle approximate any continuous operator, it restricts both subnetworks to have a single hidden layer and identical structure. 
Lu et al.~\cite{Lu2021} removed these constraints and proved an analogous approximation guarantee for arbitrary (potentially deep and asymmetric) branch and trunk architectures. 
We summarize their extension as Theorem~\ref{univapproxop2} below.
}

\begin{theorem}[\textbf{Generalized Universal Approximation Theorem for Operators},~\cite{Lu2021}]
\label{univapproxop2}
Suppose $X$ is Banach space, $K_1 \subset X$, $K_2 \subset \mathbb{R}^d$ are two compact sets, respectively, $V$ is compact in $C(K_1)$, and $G: V \rightarrow C(K_2)$ is a nonlinear continuous operator. Then, for any $\epsilon >0$, there exist positive integers $m, p,$ continuous vector functions $\pmb{g}: \mathbb{R}^m \to \mathbb{R}^p$, $\pmb{f}: \mathbb{R} \to \mathbb{R}^p$, and $x_1, \hdots, x_m \in K_1$, such that

$$\left|G(u)(y) - \langle \underbrace{\pmb{g}(u(x_1), u(x_2), ... u(x_m))}_{\text{Branch}}, \underbrace{\pmb{f}(y)}_{\text{Trunk}} \rangle\right| < \epsilon, $$
holds for all $u \in V$ and $y \in K_2$, where $\langle \cdot, \cdot \rangle$ denotes the dot product in $\mathbb{R}^p$. Furthermore, the functions $\pmb{g}$ and $\pmb{f}$ can be chosen as diverse classes of neural networks, which satisfy the classical universal approximation theorem of functions, for examples, (stacked/unstacked) fully connected neural networks, residual neural networks and convolutional neural networks.
\end{theorem}

{\color{black}
The generalization in Theorem \ref{univapproxop2} lets the branch and trunk networks differ in depth and width, provided they both output a $
p-$dimensional vector.  Assume that the output of the branch and the trunk networks are the vectors $[\beta_0\left(\pmb{u}\right),\hdots,\beta_p\left(\pmb{u}\right)]^T$ and $[\tau_0(y),\hdots,\tau_p(y)]^T$ respectively, where $\pmb{u} = \{u(x_i)\}^m_{i=1}$. Then, the output of the DeepONet; the approximate operator of $G(u)$, is defined as:
\begin{equation}
    \label{output_DeepONet}
    \hat{G}(u)(y) = \sum_{k=1}^p\beta_k(\pmb{u})\tau_k(y).
\end{equation}
Alternatively, one can add bias to each $\beta_k$ of the brand network and to the last stage of the network. Even though bias is not required in the statement of Theorem~\ref{univapproxop2}, adding bias may increase the performance by reducing the generalization error~\cite{Lu2021}. As a result, one can replace the approximate operator $\hat{G}(u)$ in \eqref{output_DeepONet} by
\begin{equation}
    \label{output_DeepONet_bias}
    \tilde{G}(u)(y) = \sum_{k=1}^p\tilde{\beta}_k(\pmb{u})\tau_k(y)+b_0,
\end{equation}
where $\tilde{\beta}_k = \beta_k + b_k, \; k=1, \hdots, p$ and $\{b_k\}^p_0$ is the set of bias.}

\begin{figure}[h]
    \centering
    \includegraphics[scale=0.35]{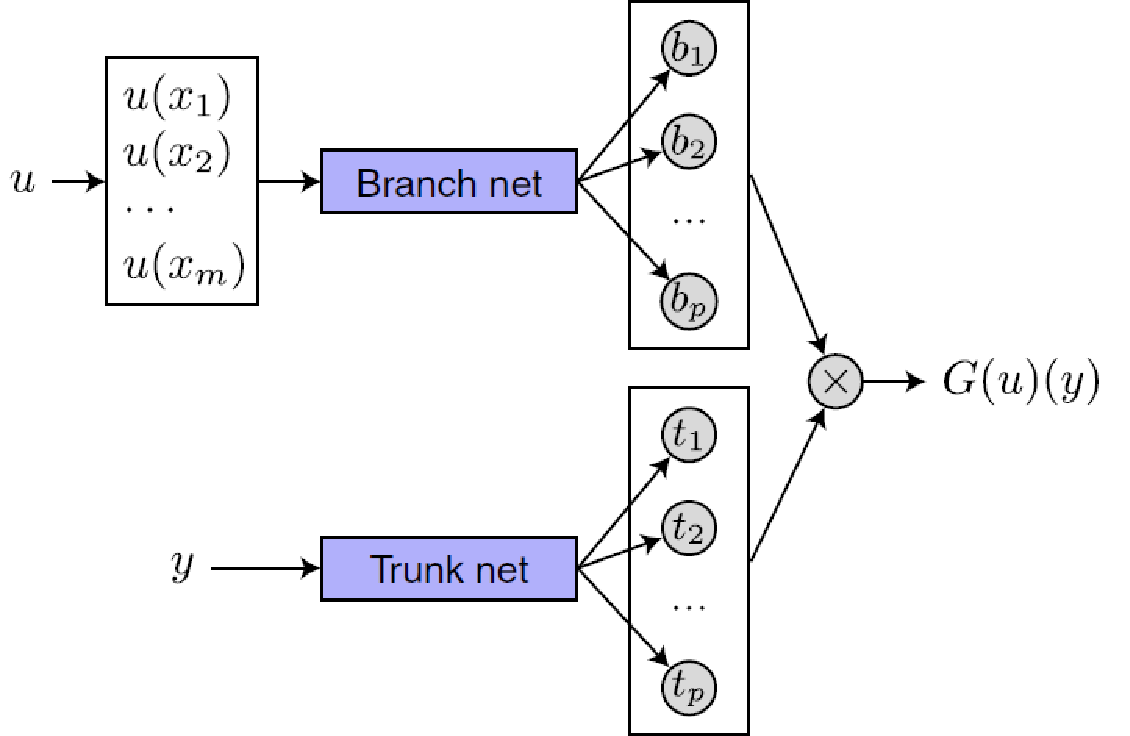}
    \caption{Vanilla DeepONet Architecture~\cite{Lu2019}. The branch networks learns input function $u(x)$ evaluated at $m$ sensors, while the trunk net learns the domain $y$ of operator output function $G(u)(y)$.}
    \label{fig:DeepONet Diagram}
\end{figure}
{\color{black}
In operator‐learning applications, we assemble the DeepONet training and testing data by forming the Cartesian product of two discrete input sets for the branch network and the trunk network. More specifically, for the branch inputs, a collection of $n$ input functions $\{u_i\}^n_{i=1}$, each sampled at $m$ fixed sensor locations $\{x_j\}^m_{j=1}$, which results in the feature vectors:
\begin{align*}
    \pmb{u}_i = \left[u_i(x_1), \hdots, u_i(x_m)\right]^T \in \mathbb{R}^m, \; i=1, \hdots, n.
\end{align*}
Regarding the inputs for the trunk architecture, we consider a set of  evaluation points $\{y_k\}^d_{k=1}$ in the operator output domain. By pairing each branch vector $\pmb{u}_i$ with each evaluation point $y_k$, we obtain a total of $n \times d$ samples:
\begin{align*}
    \left(\pmb{u}_i, \; y_k\right) \mapsto G(u_i)(y_k), \; i=1, \hdots, n, k=1, \hdots, d.
\end{align*}
This construction ensures flexibility, as the sensor grid $\{x_j\}$ fed into the branch net and the evaluation points $\{y_k\}$ used by the trunk net may differ in both dimensionality and resolution, which allows diverse sampling strategies in high-dimensional operator learning tasks.

Despite their capabilities in learning the operators between spaces, vanilla DeepONet yields fully deterministic outputs and cannot capture the intrinsic noise inherent in solutions of stochastic differential equations (SDEs).  In the next section, we give a concise overview of the Stochastic Neural Networks (SNNs), which is a probabilistic operator learning framework that integrates an SDE‐based diffusion term to recover the randomness occurring in stochastic operators.
}

\section{Stochastic Neural Network}\label{SNN}
{\color{black}
In this section, we review the mathematical formulation of Stochastic Neural Networks (SNNs) introduced in \cite{Bao2022}, and summarize the sample‐wise backpropagation algorithm for training them from \cite{Archibald2024}. As an extension of the Neural-ODE architecture, the class of SNNs in~\cite{Bao2022} was formulated by stochastic differential equations, which describe stochastic forward propagation of deep neural networks (DNNs). The training procedure of the SNNs is then designed as a stochastic optimal control problem, which is solved via a generalized stochastic gradient descent algorithm.

}

{\color{black}

\subsection{Mathematical formulation}

The SNN strucutre that we consider in this work is given by
\begin{equation}
\label{SNN_1layer}
    \begin{array}{l}
    A_{n+1} = A_n + h\mu(A_n, \theta_n)+\sqrt{h}\sigma(\theta_n) \omega_n, \; \; n=0, 1, \hdots, N-1,
    \end{array}
\end{equation}
where $A_n := \left[a^1, a^2, \hdots, a^L\right] \in \mathbb{R}^{L \times d}$ denotes the vector that contains all the $L$ neurons at the $n-$th layer in the DNN, $h$ is a fixed positive constant the stabilizes the network, $\theta_n$ represents the trainable neural network parameters, such as weights and biases, that determine the output of the neural network, $\{\omega_n\}_n:= \{\omega_n\}^{N-1}_{n=0}$ is a sequence of i.i.d standard Gaussian random variables, $\sigma$ is a coefficient function that determines the size of the uncertainty in the DNN, and $\mu$ is the pre-chosen activation function. We point out that the terms $\{\sigma(\theta_n)\omega_n\}$ allow the SNN to produce random output reflecting the stochastic behaviors of the target model~\cite{Archibald2024, Bao2022}. Unlike Bayesian neural networks (BNNs), which treat the weights $\theta$ of a standard DNN as random variables and use Bayesian inference to approximate their posterior distribution, the SNN framework introduces uncertainty by injecting a learned noise term directly into the network dynamics.

If we choose a positive constant $T$ as a pseudo terminal time and let $N\rightarrow \infty$ or $h \rightarrow 0$ with $h = T/N$, the dynamics of the SNN~\eqref{SNN_1layer} becomes the stochastic differential equation, which, in the integral form, is given by
\begin{equation}
\label{SNN_continuous}
    \begin{array}{l}
    A_T = A_0 + \mathlarger{\int}_0^T \mu( A_t, \theta_t)dt + \mathlarger{\int}_0^T \sigma(\theta_t) dW_t,
    \end{array}
\end{equation}
where $W:= \left\{W_t\right\}_{0 \leq t \leq T}$ is a standard Brownian motion corresponding to the i.i.d. Gaussian random variable sequence $\{w_n\}_n$ in~\eqref{SNN_1layer}, $\mathlarger{\int}^T_0\sigma(\theta_t)dW_t$ is an It\^{o} integral, and $A_T$ corresponds to the output of the SNN. 

}



{\color{black}
In this work, we shall treat the training procedure for deep
learning as a stochastic optimal control problem and the parameter $\theta_t$ in~\eqref{SNN_continuous} as a control process. This approach is supported by the Stochastic Maximum Principle (SMP), and then translated into an implementable architecture with diverse applications~\cite{Archibald2024}. An alternative formulation based on the Dynamic Programming Principle (DPP) leads to Hamilton–Jacobi–Bellman (HJB) equations.  However, the high dimensionality of neural network parameter spaces renders the direct solution of HJB equations computationally intractable.  Consequently, most SNN architectures rely on SMP rather than DPP to ensure scalability and numerical tractability in learning stochastic operators.

To this end, let $\Gamma$ be the random variable that generates the training data to be compared with $A_T$, we define the cost functional $J$ for the optimal control problem as
\begin{align*}
    \begin{array}{l}
   J(\theta)= E \left[{\Phi(A_T, \Gamma)} + \mathlarger{\int}_0^T r(A_t, \theta_t) dt \right],
    \end{array}
\end{align*}
where $\Phi\left(A_{T}, \Gamma\right):= \vert\vert A_T-\Gamma\vert\vert_{\text{loss}}$ is a loss function corresponding to a loss error norm $\vert\vert \cdot \vert\vert_{\text{loss}}$, and the integral $\mathlarger{\int}_0^T r(A_t, \theta_t) dt$ represents the running cost in a control problem.
The goal of the deep learning is to solve the stochastic optimal control problem, i.e., find the optimal control $\theta^*$ such that
\begin{equation}
    \begin{array}{l}
    \label{optimal_control}
    J(\theta^*)=\inf\limits_{\theta \in \Theta}J(\theta),
    \end{array}
\end{equation}
where $\Theta$ is an admissible control set.
}

We assume that the control space $\Theta$ is convex and that the control $\theta_t \in \Theta$ is an admissible control~\cite{Bao2020a}. An admissible control is one that is both square integrable: $\int_{-\infty}^{\infty}|\theta_s|^2 ds < \infty$ and adapted to the filtration $\mathbb{F}$ generated by $W_s$. Since $\theta_t$ is the set of neural network parameters, we think of $\Theta = \mathbb{R}$ or that these weights and biases might become any real number, thus ensuring that $\Theta$ is convex.

The stochastic maximum principle (SMP) allows us to solve the optimal control problem by minimizing the Hamiltonian defined by:

$$H(a, b, c, \theta) = r(a,\theta) + b\mu(a,\theta) + c \sigma(\theta). $$


\begin{theorem}[Stochastic Maximum Principle (SMP)~\cite{Andersson2009}]
\label{theorem: SMP}
    Let $\theta \in \Theta$ such that $E[|\theta|^2]<\infty$ and $A^*$ be the optimally controlled process with optimal control $\theta^*$. Then there exists a pair of adjoint processes $(B^{*}, C^{*})$ s.t.:
    \begin{equation}
    \label{FB_SDEs}
        \begin{array}{l}
        {\color{black} dA^{*}_t =  \mu(A^{*}_t,\theta_t)dt+\sigma(\theta_t)dW_t}, \vspace{0.1cm} \\
        {\color{black} dB^{*}_t= -(r_a(A^{*}_t, \theta^*_t) +\mu_a(A^{*}_t, \theta^*_t)^TB^{*}_t + \sigma_a( \theta^*_t)^TC^{*}_t)dt -C^{*}_tdW_t},
        \end{array}
    \end{equation}
with 
$B^{*}_T=\Phi_a(A^{*}_T)$ and $C^{*}_t=\sigma_t\nabla_aB^{*}_t$.
    
Additionally,

$$H(A^*_t, B^*_t, C^*_t, \theta^*_t) = \inf_{\theta \in \Theta}H(A^*_t, B^*_t, C^*_t, \theta_t).$$
\end{theorem}

In order to determine the optimal controls $\theta$, one can derive the following gradient process~\cite{Ma1999, Yang2021}:

$$\nabla_{\theta}J(\theta)|_t = E\left[\mu_\theta^T(t)B_t + \sigma_\theta^T(t)C_t+ r_\theta(t)\right].$$
Therefore, the gradient of the loss function is the expectation of the gradient of the Hamiltonian of SMP. Now, suppose that $\theta$ are the parameters of a neural network. It is clear then that one can replace the gradient of the original loss function when performing stochastic gradient descent with the expected gradient of the Hamiltonian, giving the following update on the SNN weights~\cite{Bao2020a}:
\begin{equation}
\begin{array}{l}
\theta^{i+1}_t =\mathcal{P}_{\Theta}( \theta^i_t + \alpha E\left[\nabla_\theta H(A_t,B_t,C_t,\theta)\right]),
\end{array}
\end{equation}
where $\alpha$ is the step size or learning rate, and $\mathcal{P}_{\Theta}$ is the projection operator onto the set of admissible controls. Then, as is well established for SGD, one takes single sample approximations $A^i_t$, $B_t^i$, $C^i_t$ of $A_t$, $B_t$, and $C_t$ at training iteration $i$, resulting in the final update~\cite{Bottou2018}:
\vspace{-0.1cm}
\begin{equation}   
\label{stochastic_gradient_descent}
    \begin{array}{l}
       \theta^{i+1}_t =\mathcal{P}_{\Theta}( \theta^i_t + \alpha \left[r_\theta(A_t^i;\theta^i_t) + \mu_\theta^T(A_t^i; \theta^i)B^i_t + \sigma_\theta^T(\theta_t^i)C_t^i\right]).
    \end{array}
\end{equation}
In~\cite{Archibald2024}, the convergence of this SGD iteration was proven under the standard smoothness assumptions for SMP. {\color{black} Numerical implementation of the stochastic gradient descent scheme~\eqref{stochastic_gradient_descent} requires
numerical approximations of the forward and backward SDEs in~\eqref{FB_SDEs}. In the next section, we briefly introduce the numerical schemes for approximately solving these equations.}

\subsection{Numerical Approximation for the forward and backward SDEs}
Given that the running cost $r(A_t,\theta)$ is typically not included in most loss functions utilized in machine learning, the main challenge to implement the SNN SGD iteration is the numerical approximation of the backward SDE solutions $B^i_t$ and $C^i_t$. We first define the partitioning scheme:

$$\Pi_N = \{t_n, 0=t_0<t_1<...<t_N=T\},$$ 
where $\Pi_N$ divides the pseudo-time $[0,T]$ upon which SDE $A_t$ is defined into $N$ uniform intervals. For SNN, $N$ represents the number of layers and we always take terminal pseudo-time $T=1$. Hence, each interval is of size $h=\frac{T}{N}=\frac{1}{N}$. The partition $\Pi_N$ is then employed to split the forward and backward SDEs into $N+1$ parts to get
$$A_{t_{n+1}} = A_{t_n} + \int_{t_n}^{t_{n+1}} \mu(A_s, \theta_s)ds + \int_{t_n}^{t_{n+1}} \sigma( \theta_s) dWs, \; n=0, \hdots, N-1,$$
and
\begin{align*}
    \begin{array}{l}
    B_{t_n}=B_{t_{n+1}} +\mathlarger{\int}_{t_n}^{t_{n+1}}(r_a(A_s, \theta_s) +\mu_a(A_s, \theta_s)^TB_s + \sigma_a(u_s)^TC_s)ds -\mathlarger{\int}_{t_n}^{t_{n+1}}C_sdW_s, \vspace{0.1cm} \\
    \hspace{9cm} n = N-1, \hdots, 0.
    \end{array}
\end{align*}
The state $A_t$ can then be approximated by the Euler-Maruyama scheme for discretizing SDEs ~\cite{Bao2020a}:
$$A^i_{t_{n+1}} = A^i_{t_n} + \mu(A^i_{t_n}, \theta^i_{t_n})h + \sigma( \theta^i_{t_n}) \Delta W_n, \; n=0, \hdots, N-1.$$
This acts as the forward pass of the stochastic neural network.

For the adjoint equation $B_t$, a problem arises from that fact that while the equation flows backwards in time, we also assumed that $B_t$ and $C_t$ are both adapted to the filtration $\mathcal{F}^{W}_t$, where $\mathcal{F}^{W}_t:= \sigma\left(W_s, 0 \leq s \leq t\right)$ is the $\sigma-$ algebra generated by $\{W_s\}_{0 \leq s \leq t}$. Hence, despite needing to calculate $B_{t_n}$ from $B_{t_{n+1}}$, we only have the information up to time $t_n$, and therefore we must calculate $B_{t_n}$ under conditional expectation with respect to the filtration $\mathcal{F}^{W}_{t_n}$. This results in, for $n=N-1, \hdots, 0,$
\begin{align*}
    \begin{array}{l}
B_{t_n}=E[B_{t_{n+1}}|\mathcal{F}^{W}_{t_n}] +\mathlarger{\int}_{t_n}^{t_{n+1}}E\left[(r_a(A_s, \theta_s) +\mu_a(A_s, \theta_s)^TB_s + \sigma_a(\theta_s)^TC_s)|\mathcal{F}_{t_n}\right]ds,
    \end{array}
\end{align*}
as $E\left[\mathlarger{\int}_{t_n}^{t_{n+1}} C_s dW_s\right] =0$ by the definition of the stochastic integral~\cite{Bao2020a}. Similarly, we can use the Ito isometry that $(dW_t)^2=dt$ under expectation to get~\cite{Bao2020a, bao2016first, Bao_DA_BSDE, Bjork2019}:

$$C_{t_{n}}=\frac{E[B_{t_{n+1}}\Delta W_n|\mathcal{F}^{W}_{t_n}]}{h}.$$
As with SGD, we can then use a single sample as an unbiased estimator of the conditional expectation to obtain:
\begin{equation}
    \begin{array}{l}
    B^i_{t_{n}} = B^i_{t_{n+1}} + h\left[\nabla_{a} H( A^i_{t_{n+1}}, B^i_{t_{n+1}}, C^i_{t_{n}}, \theta_{t_n})\right], \vspace{0.1cm} \\
    C^i_{t_{n}} =\dfrac{B^i_{t_{n+1}}\epsilon_{t_n}}{\sqrt{h}}, 
    \end{array}, \; n = N-1, \hdots, 0,
\end{equation}
where $\epsilon_{t_n}$ is drawn from the standard normal distribution, as standard Brownian Motion $\sim N(0, t)$~\cite{Bao2018}. These two formulae act as the first step of the backward pass when training SNN, and are used to compute the Hamiltonian, whose gradient is in turn used to update the weights $\theta$.

{\color{black} Finally, we integrate the SNN training procedure into the DeepONet architecture to obtain a unified stochastic operator‐learning framework, which we name the Stochastic Operator Network.}

\section{Implementation of Stochastic Operator Network}\label{SON}


Stochastic Operator Network (SON) replaces the branch network of DeepONet with the Stochastic Neural Network, allowing it to learn noisy operators such as stochastic ODEs through uncertainty quantification. {\color{black} Although the specific drift and diffusion sub‐networks in the Euler‐SDE scheme vary with each application, SON consistently trains both sets of parameters to capture the pointwise mean and variance of the operator’s output.}

Suppose we want to approximate the following noisy operator
$G(u)(y, \epsilon)$
where {\color{black} $u(x)$ is the input function and} $\epsilon$ is a random noise for each $y$ that obscures the operator output function in $\mathcal{Y}$. To solve this problem, we first define the forward neural SDE for $N$ pseudo-time steps in $t \in [0,T]$ where $T=1$:
$$A^i_{t_{n+1}} = A^i_{t_n} + \mu(A^i_{t_n}, \theta^i_{t_n})\Delta t + \sigma(A^i_{t_n}, \theta^i_{t_n}) \sqrt{\Delta t} \epsilon, \; n=0, \hdots, N-1,$$
where $\Delta t=\frac{1}{N}$, and $\mu_{t_n}(\cdot, \theta)$ is produced by a neural network with trainable parameters $\theta$ at each time $t_n$. Likewise, the diffusion coefficient $\sigma(\cdot, \theta)$ can either be implemented as a simple list of trainable parameters or, like the drift, be defined by its own neural network. The branch net is therefore:
$$\beta(u; \theta_{\beta},\epsilon) = A_{T}(A_{T-\Delta t}(...A_0(u; \theta_0)...;\theta_{T-\Delta t});\theta_T).$$
Therefore, a single branch net layer of SON is one forward propagation of the Euler-Maruyama scheme by $\frac{1}{N}$ along the SNN pseudo time interval $[0,1]$. The trunk net is a feed-forward NN with $L_\tau$ layers:
$$\tau(y, \theta) = \tau_{L_\tau}(\tau_{L_\tau-1}(...\tau_0(y,\theta_0)...,\theta_{L_\tau-1}), \theta_{L_\tau}).$$
The output of both the SNN and trunk net should be vectors in $\mathbb{R}^p$. As with DeepONet, SON combines the outputs of these parallel structures together with the inner product to produce the operator output:
$$\hat{G}_{SON}(u)(y) = \sum_{k=1}^p\beta(u; \theta_{\beta}, \epsilon)\tau(y; \theta_\tau) +b_0,$$
where $b_0$ is a trainable bias parameter to improve generalization.

{\color{black}
In our framework, the uncertainty inherent to the target stochastic operator is generated during the forward pass: we encode the input function $u(x)$ into an initial latent state and then evolve that state under an SDE whose drift and diffusion networks both depend on $u$. The resulting terminal state of this SDE serves as the output for the branch network, embedding randomness into the operator mapping itself rather than adding noise to an otherwise deterministic output.

Although we integrate DeepONet’s reconstruction layer onto an SNN ResNet backbone, we cannot simply train its parameters by standard backpropagation alone. Because the ResNet corresponds to a time‐discretized SDE under stochastic‐optimal‐control, weight updates must follow the Hamiltonian‐gradient conditions of the Stochastic Maximum Principle defined in Section~\ref{SNN}. In practice, SON performs one outer DeepONet evaluation per SNN step and then applies the adjoint–BSDE update to the drift and diffusion weights. Therefore, SON is more mechanically akin to adding DeepONet as the final layer of stochastic neural network. The full training loop is summarized in Algorithm \ref{SONTraining}.
}


\begin{algorithm}\label{alg: SONalg}
    \caption{SON Training Procedure}\label{SONTraining}
    For each epoch $i$ in $i=0,...,I$:
    \begin{enumerate}
        \item Randomly draw a training pair \((u,y)\) from the dataset, where the input function \(u\) is discretized at grid points \(\{x_0,\dots,x_m\}\) into the vector
    \[
      \pmb{u} = (u_0,\dots,u_m)\;=\;(u(x_0),\dots,u(x_m)).
    \]
    \item Let $A_{t_0} = \pmb{u}$. For $n=0, \hdots, N-1$,
  \begin{enumerate}
      \item Compute $\mu(A_{t_n}, \theta)$ and $\sigma(A_{t_n}, \theta)$ by either constant of DNN layers.
      \item $A_{t_{n+1}} \gets A_{t_n} + \mu(A_{t_n}; \theta_{t_n})\Delta t + \sigma(A_{t_n}; \theta_{t_n}) \sqrt{\Delta t} \epsilon.$
  \end{enumerate}
       \item  $\hat{y} \gets Trunk(y).$
        \item $\hat{G}_{SON}(u)(y) \gets\langle \hat{y}, u^i_{t_N} \rangle$
        \item $B_{t_N}= \nabla_u \Phi(\hat{G}_{SON}(u)(y))$.
        \item Solve backward for each $B_{t_n}$ and $C_{t_n}$ by computing Hamiltonian and taking $\nabla_u H(\cdot)$
        \item $$\theta^{i+1}_t = \theta^i_t + \alpha \nabla_\theta H(A_{t_n},B_{t_n},C_{t_n};\theta).$$
    \end{enumerate}
\end{algorithm}

It is important to note that the dimension of $u$ during SNN forward propagation must remain consistent, so any projection of data must be done in layers before or after the Euler-Maruyama iteration~\cite{Archibald2024}. If we have projections, then we project from the space $C(K_1)$ where our input functions $u(x)$ exist down to the space where we want to propagate the SDE, run the SDE forward, then project down to the space where the inner product takes place. Calculating $B_t$ for projections while backpropagating is akin to setting $C_t =0$, since no additional noise is added by the network during the forward pass. The same is also true for the final layer, where the inner product between the trunk and SNN is taken. This is therefore the same as backpropagation for a single step of a neural ODE~\cite{Chen2018}.


\section{Numerical Experiments} \label{EXP}
\subsection{Data Generation}

To construct the train and test datasets, input functions $u(x)$ were sampled from a mean zero Gaussian random field (GRF) using the RBF covariance kernel with length scale paramter $l = 0.2$, as in~\cite{Lu2019}. Values $y$ were also randomly sampled from the operator output domain and then the cartesian product of these two sets was formed. A single training example is therefore $(y, u(x))$ where $y \in \mathbb{R}^n$, $u(x) \in \mathbb{R}^m$ is an array of input function values, $n$ is the dimension of the domain of operator output $G(u)(y)$, and $m$ is the number of sensors, $x$.  The correct operator output values $G(u)(y)$ were then calculated numerically using explicit Runge Kutta (RK45) in JAX. 

{\color{black} In this section, we carry out five numerical experiments to demonstrate the performance of our proposed method in learning the stochastic operators. We then compare the results obtained by our method with those produced by the vanilla DeepONet.}

\subsection{Antiderivative Operator with Noise}\label{ex1}
{\color{black} We first consider the following antiderivative operator with noise:}

 $$G: u(x) \to s=s(0) +  \int_0^y u(\tau) d\tau + \alpha\epsilon,$$
 with the initial condition $s(0)=0$ and $\epsilon \sim N(0,1)$ noise scaled by constant $\alpha =0.1$. {\color{black} We aim to train a neural network to approximate the target operator and provide an estimation to the scaling constant $\alpha$.}
 
 One hundred training $u(x)$ functions were sampled from GRF, with 100 sensors uniformly partitioning $x \in [0,5]$, and uniformly randomly sampled 100 training points $t$ from the operator output domain $t \in[0,5]$. This gives $100 \times 100 = 10,000$ training examples. For testing, this was $1000\text{ } u(x) \times 1000\text{ } t = 1,000,000$ examples, with the $t$ uniformly partitioning $[0,5]$. 

 For the trunk net, a 2 layer dense architecture with ReLU activation was implemented. For the SNN, 6 layers were used (i.e. $\Delta n =\frac{1}{6}$ on $[0,1]$ pseudo time). The drift $\mu$ was a 3 layer dense neural network with ReLU activation for each pseudo time step, while diffusion $\sigma$ was a single trainable parameter at each pseudo time step initialized via $N(0,1)$ normal distribution. Since dimension must be preserved during SNN forward propagation, $\mu$ had 100 neurons per layer as each $u(x)$ had 100 sensors. This gives the following form for the SNN forward discretized SDE:
 $$u_{n+1} = u_n+ \mu(u_n; \theta)\Delta n + \sigma(\theta)\sqrt{\Delta n}\epsilon, \; n=0, \hdots, N-1,$$
where $\theta$ are trainable weights. 

SON trained for 2000 epochs with a 0.001 learning rate using ADAM and without batches. Mean squared error was the loss function for the initial BSDE backpropagation. One key thing to note is that while the loss maintains a generally convergent trajectory during operator learning, it can spike wildly due to the architecture and complexity of the problem. Therefore, a scheduler was used, multiplying the learning rate by 0.9 every 500 epochs after 1000. 
     \begin{figure}[h!]
        \begin{minipage}{0.47\textwidth}
            \centering
        \includegraphics[scale=0.24]{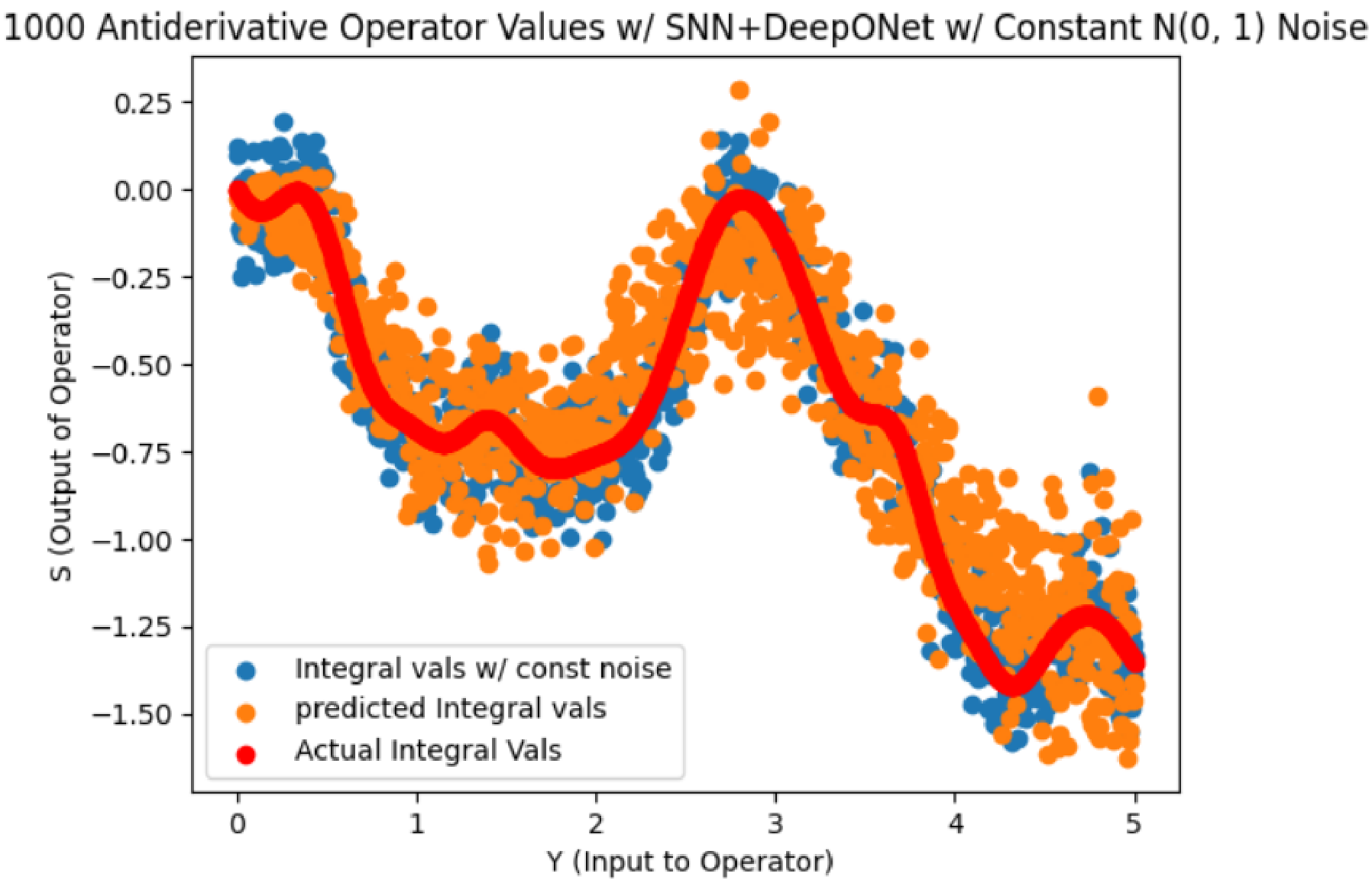}
        \end{minipage}
        \begin{minipage}{0.47\textwidth}
             \centering
        \includegraphics[scale=0.24]{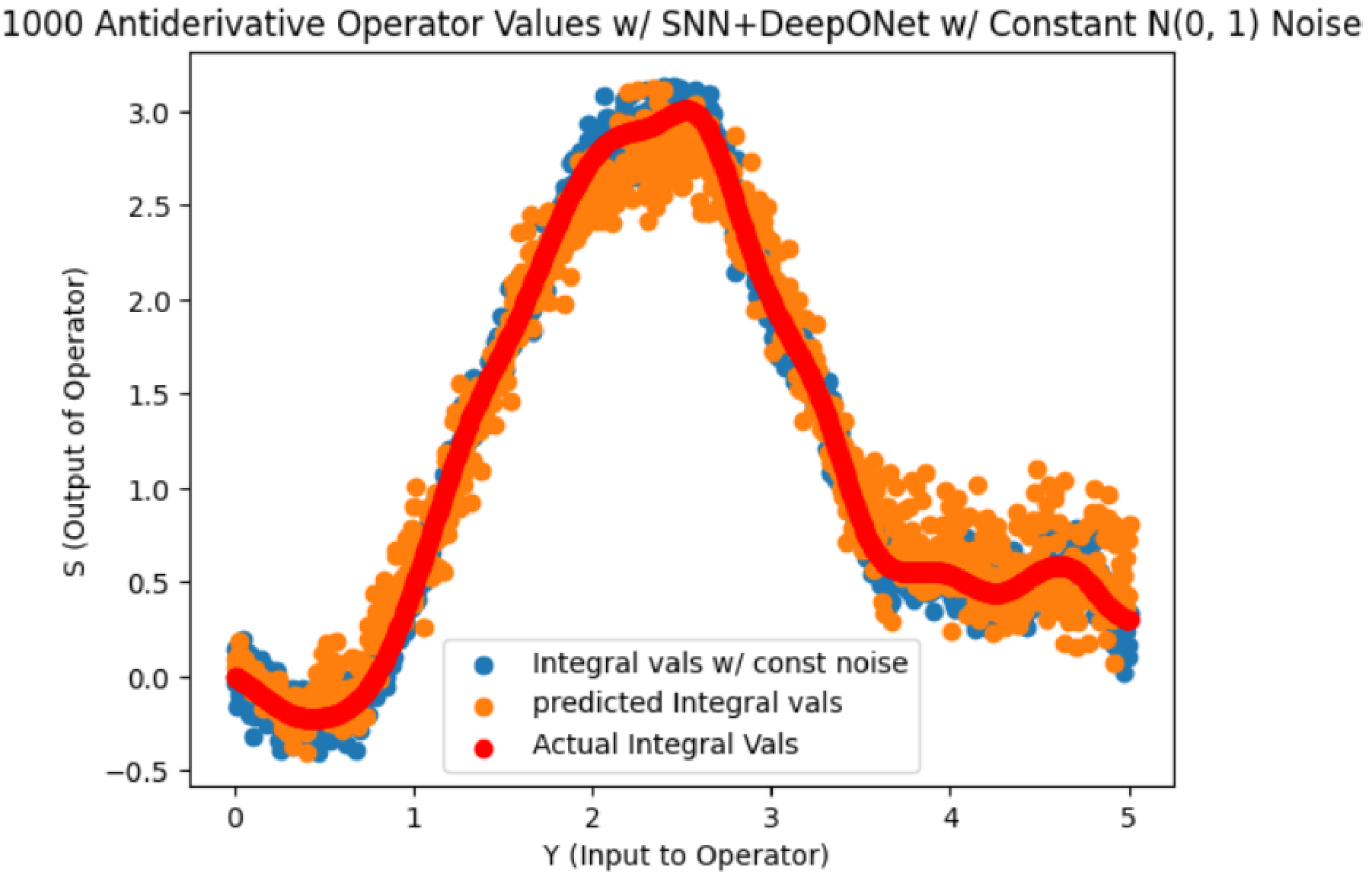}
        \end{minipage}
        
        \begin{minipage}{0.47\textwidth}
            \centering
        \includegraphics[scale=0.24]{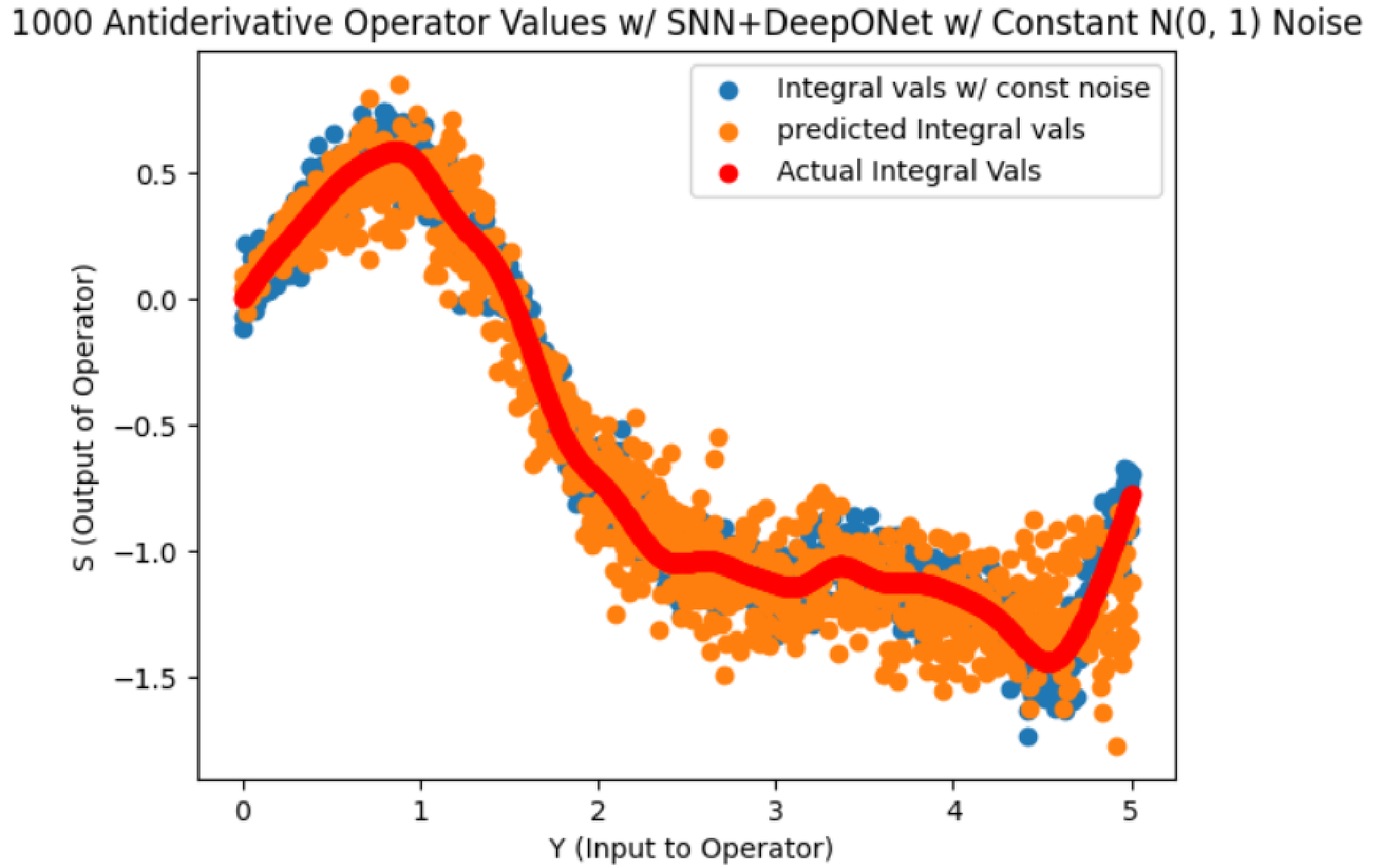}
        \end{minipage}
        \begin{minipage}{0.47\textwidth}
             \centering
        \includegraphics[scale=0.24]{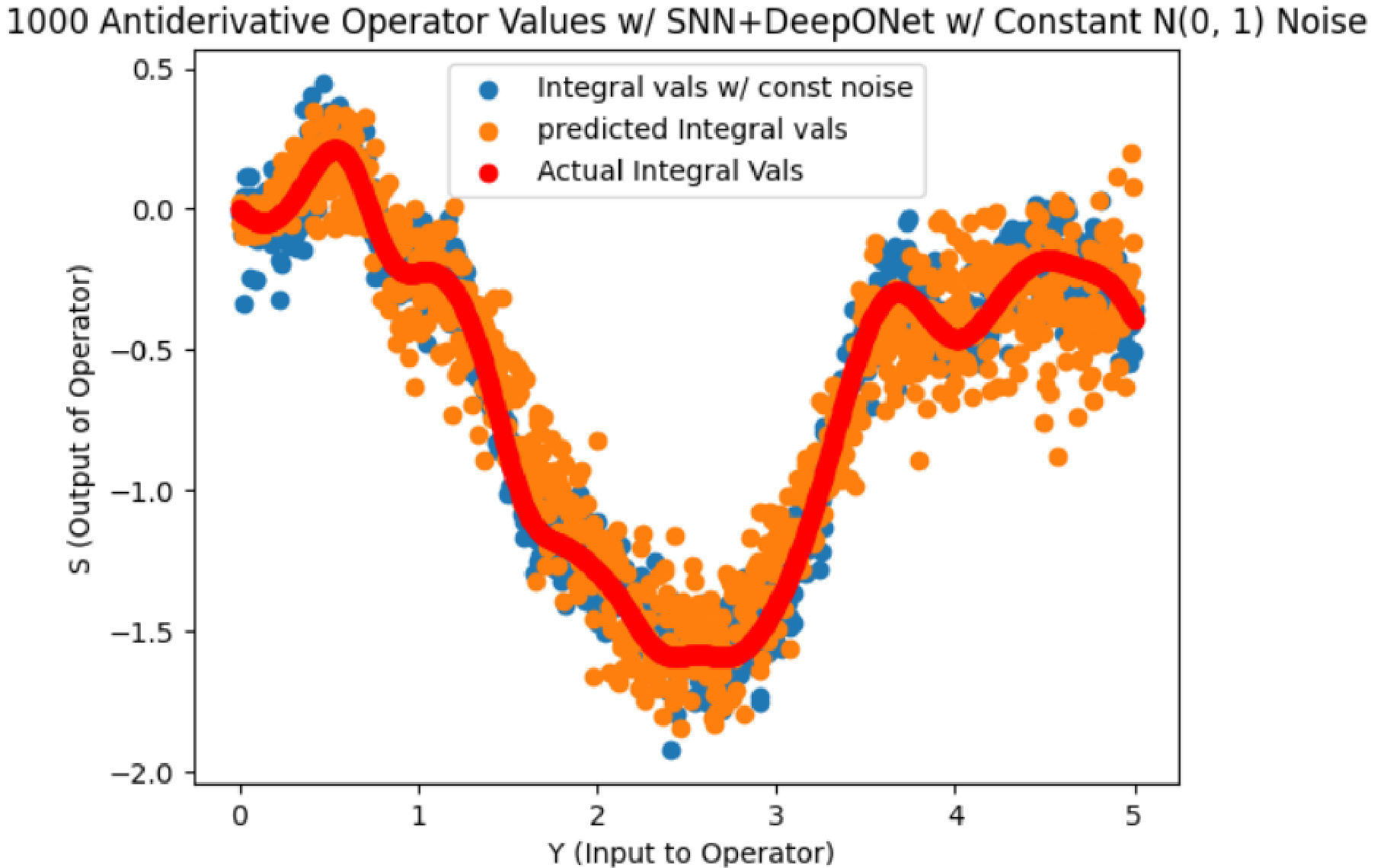}
        \end{minipage}
        \caption{Antiderivative operator applied to four random test $u(x)$.}
        \label{fig:integral_plots}
        \vspace{-0.4cm}
    \end{figure}

    \begin{figure}[h!]
        \centering
        \includegraphics[scale=0.3]{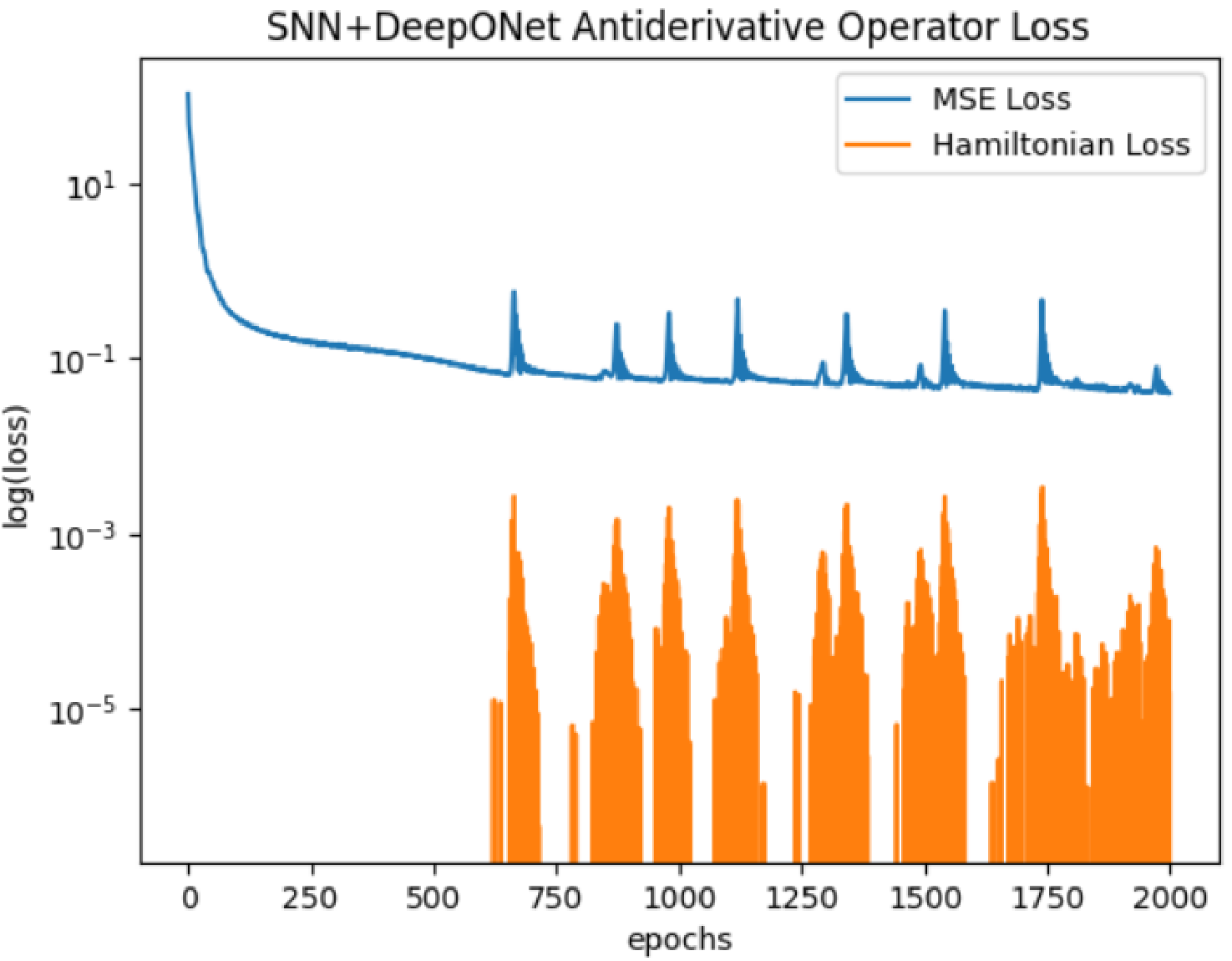}
        \caption{Noisy antiderivative operator loss over 2000 epochs.}
        \label{fig:integral loss}
        \vspace{-0.3cm}
    \end{figure}
    
    Figure \ref{fig:integral_plots} shows the integrals of four random $u(x)$ functions sampled from the test set. Red indicates the true integral values, blue indicates the anitderivative operator values with $N(0,1)$ noise scaled by 0.1 added randomly at each point, and orange indicates the SON predictions. As Figure \ref{fig:integral_plots} shows, SON is able to capture both the mean and variation of the operator output with a high degree of accuracy. The final MSE after training was $\approx 0.04$, as seen in Figure \ref{fig:integral loss}, well within the level of scaled noise added to the operator output. Since the Hamiltonian loss is not restricted to be positive, the plot only shows instances where this loss is nonnegative under the log scaling.

    In order to assess that SON has correctly quantified the uncertainty, every testing example was predicted by SON 100 times and the standard deviation calculated for each. This was then averaged over all testing examples (for both every U and every y). If SON is able to recover the scaling factor $\alpha =0.1$, which was multiplied by the noise added to the data, then it has successfully quantified the uncertainty through the diffusion $\sigma$. As Table \ref{tab:Antideriv Noise} shows, the recovered estimates are very close to the true scaling factor for three different trainings of SON on this problem. 

    \begin{table}[h!]
        \centering
        \begin{tabular}{|c|c|c|c|}
        \hline
             &  Trial 1 & Trial 2 & Trial 3\\
             \hline
            Recovered Noise & 0.1430 & 0.1639 &0.1354\\
            \hline
        \end{tabular}
        \caption{Average point-wise stdev. for noisy antiderivative operator across multiple trainings.}
        \label{tab:Antideriv Noise}
    \end{table}

\subsection{ODE with Noise}\label{ODE}
{\color{black} We next consider the perturbed solution operator to the following ODE
$$\frac{ds(y)}{dy} = s(y)u(y),$$
with initial condition $s(0)=1$. More specifically, the target stochastic operator takes the form:
\begin{equation}
    \notag
    s(y) = s(0)e^{\int_0^y u(\tau) d\tau}+\alpha\epsilon = e^{\int_0^y u(\tau) d\tau}+\alpha\epsilon,
\end{equation}
where $\epsilon \sim N(0,1)$ is the random noise scaled by $\alpha =0.1$.
}
\begin{figure}[h!]
\begin{minipage}{0.47\textwidth}
    \centering
\includegraphics[scale=0.34]{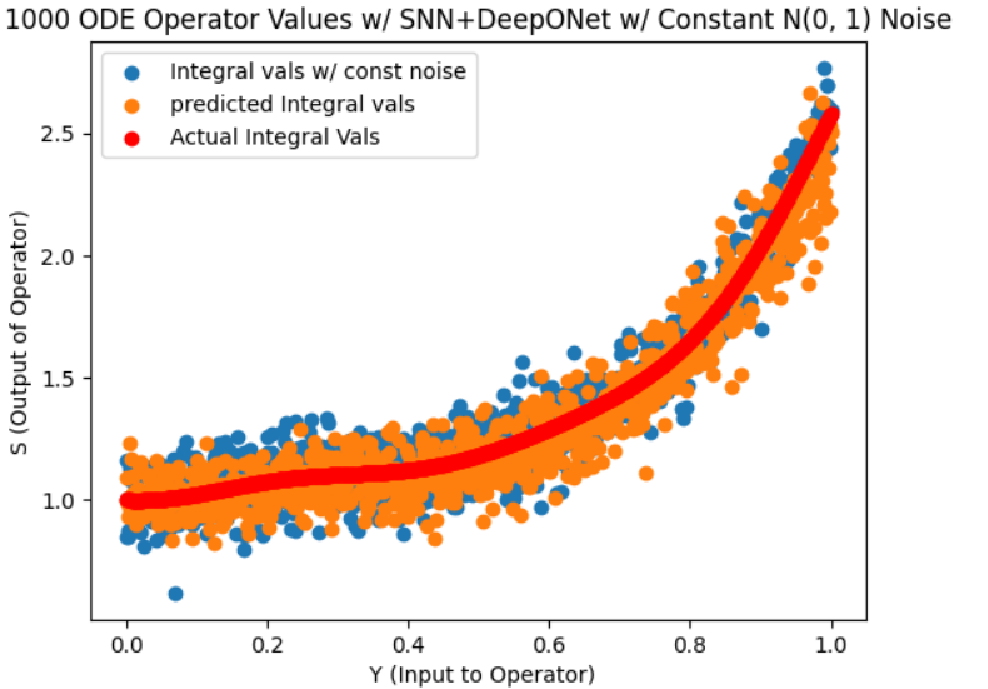}
\end{minipage}%
\begin{minipage}{0.47\textwidth}
     \centering
\includegraphics[scale=0.34]{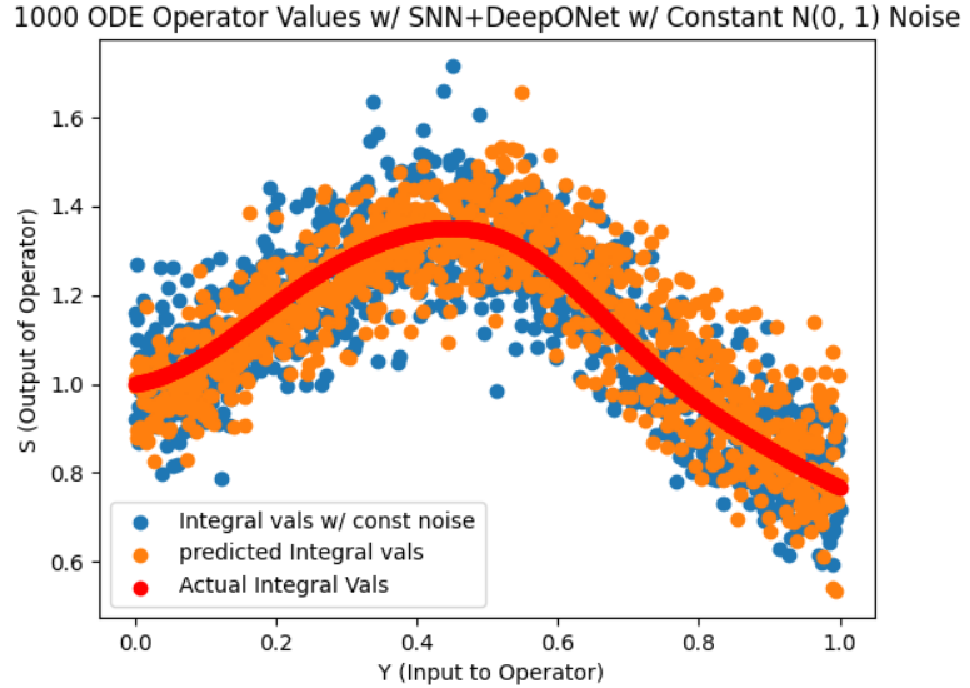}
\end{minipage}
\hfill
\begin{minipage}{0.47\textwidth}
\centering
    \includegraphics[scale=0.34]{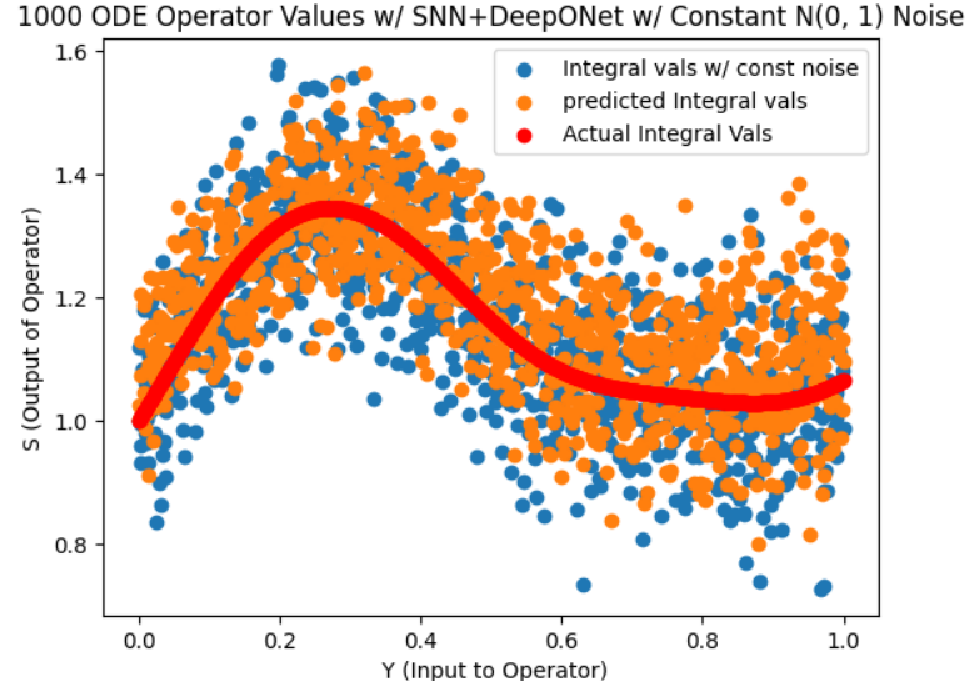}
\end{minipage}%
\begin{minipage}{0.47\textwidth}
     \centering
        \includegraphics[scale=0.34]{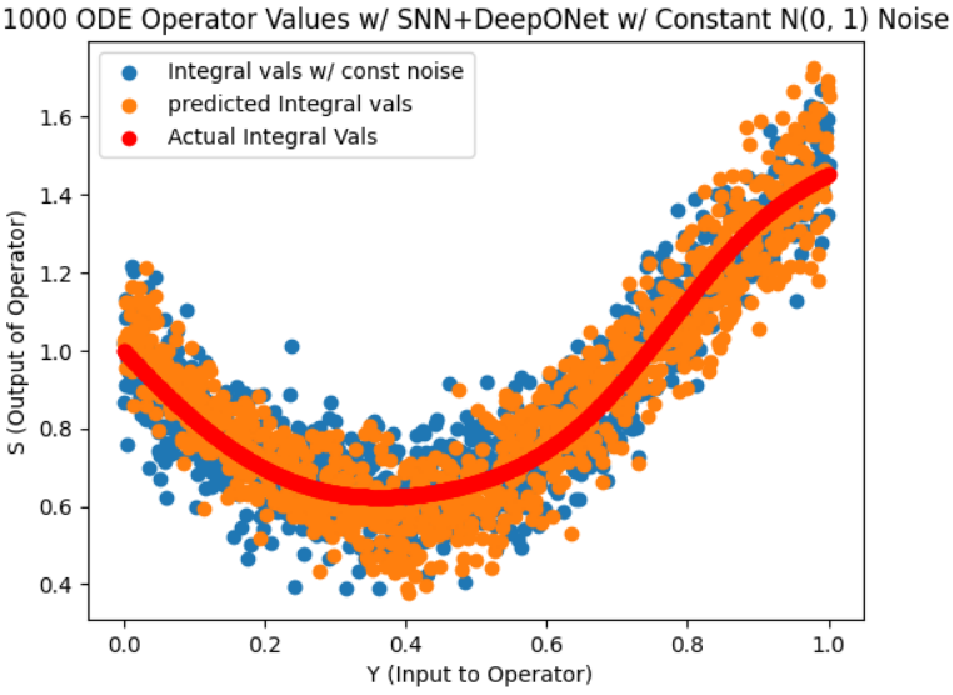}
        \end{minipage}
        \caption{\small SON learning ODE applied to four random test $u(x)$.}
        \label{fig:ODE_plots}
        \vspace{-0.3cm}
    \end{figure}
\begin{figure}[h!]
        \centering
        \includegraphics[scale=0.4]{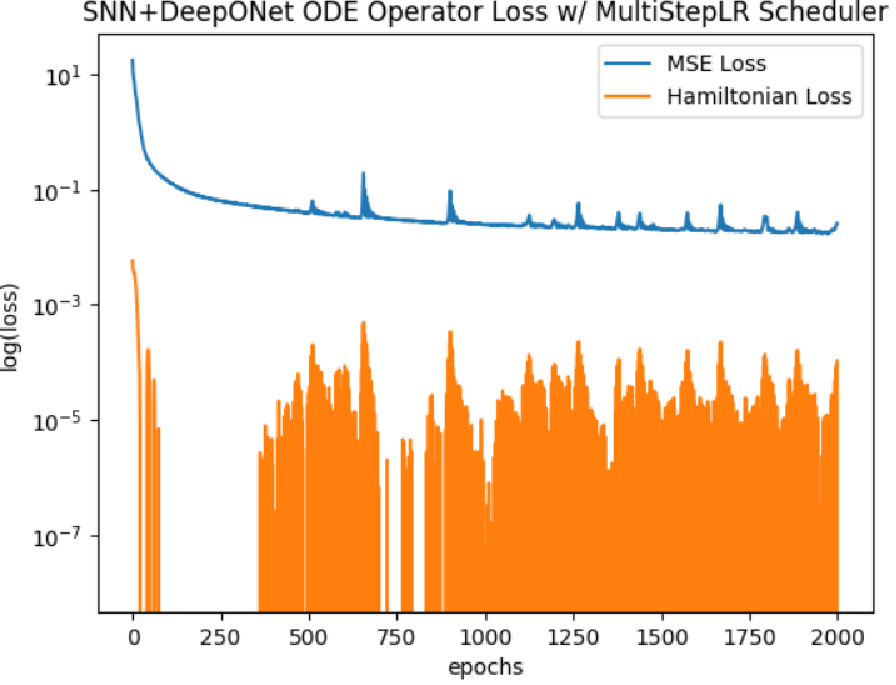}
        \caption{Noisy ODE operator loss over 2000 epochs.}
        \label{fig:ode loss}
        \vspace{-0.4cm}
\end{figure}

Training and testing data was generated as in Section \ref{ex1}, except our operator output domain was $t \in [0,1]$ to have a comparable range for all samples in the dataset. Keeping the domain $[0,5]$ resulted in difficulties learning the ODE's with much larger, outlier ranges. SON was also intialized as in Section \ref{ex1}.

As Figure \ref{fig:ODE_plots} shows, SON is able to once again able to approximate the ODE solution trajectory well for the four random test $u(x)$ shown. Final MSE was $\approx 0.0178$ (Figure \ref{fig:ode loss}) while SON was able to correctly quantify the uncertainty across two different trainings, as judged by the same standard deviation recovery test as before (Table \ref{tab:Antideriv Noise}).

\begin{table}[h!]
        \centering
        \begin{tabular}{|c|c|c|c|}
        \hline
             &  Trial 1 & Trial 2 & Trial 3 \\
             \hline
            Recovered Noise & 0.0925 & 0.0722 & 0.0985\\
            \hline
        \end{tabular}
        \caption{Average point-wise stdev. for noisy ODE operator across multiple trainings.}
        \label{tab:ODE Noise}
    \end{table}

\subsection{2D ODE System with Noise}\label{2DODE}
Our third numerical experiment is the following nonlinear 2D system of ODEs:
\begin{equation}
    \begin{array}{l}
        \dfrac{ds_1}{dy} = s_2, \vspace{0.1cm} \\
        \dfrac{ds_2}{dy} = -\sin{s_1}+u(y),          
    \end{array}
\end{equation}
where $s_0 = (s_1, s_2)=(0,0)$ and $y \in [0,1]$. {\color{black} We perturb the solution operator by adding a two-dimensional Gaussian noise vector
\begin{align*}
    \begin{array}{l}
     \pmb{\varepsilon} = \alpha\tilde{\pmb{\varepsilon}}, \; \tilde{\pmb{\varepsilon}} \sim N\left(0, \pmb{I}_2\right),\; \alpha =0.1,
    \end{array}
\end{align*}
where $\tilde{\pmb{\varepsilon}} = \left(\varepsilon_1, \varepsilon_2\right)$ with each component is sampled independently from $N(0, 1).$
}
As was done in~\cite{Lu2021}, significant changes to SON needed to be implemented during the step where the SNN and trunk nets are combined in order to produce output with increased dimensions ~\cite{Lu2021}. For this experiment, output of the trunk net was made twice as large (200 neurons) as the SNN output (100 neurons) and then split in half to create two dimensional output. Therefore, in this instance, the first 100 neurons of the trunk net effectively provide information for $s_1$ in the domain, while the second 100 neurons provide information on $s_2$. These two pieces were then combined together with the same SNN output to produce the output of the noisy system $s = G(u)(y)$.

While most other hyperparameters and architectures remained the same as the previous experiments, the depth of SNN was increased to 10 layers and the diffusion parameters were initialized with $N(0,2)$. 

\begin{figure}[h!]
    \centering
    \includegraphics[scale=0.45]{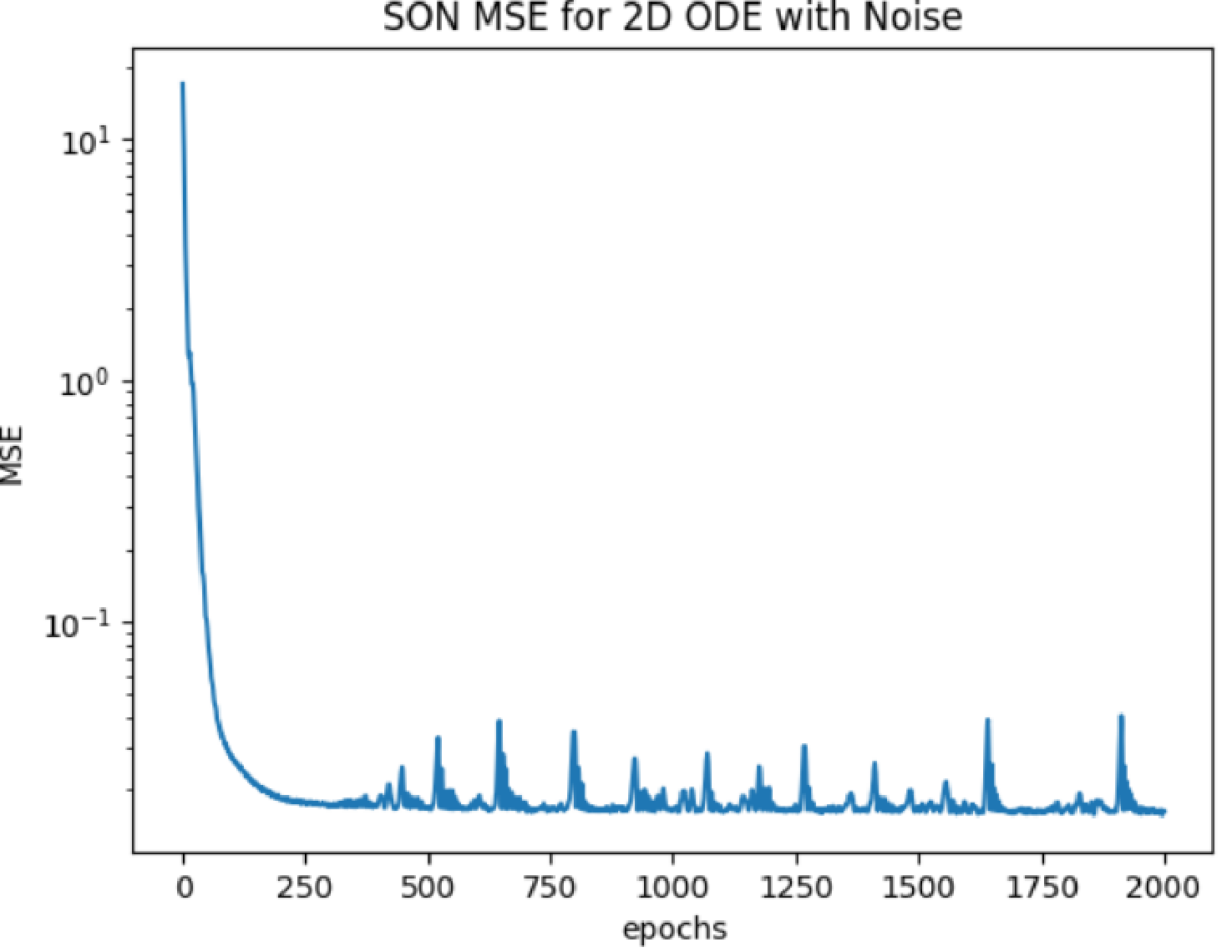}
    \caption{\small MSE loss for SON learning 2D ODE with noise.}
    \label{fig:2dodeloss}
    \vspace{-0.3cm}
\end{figure}
\begin{figure}[h!]
\begin{minipage}{0.48\textwidth}
    \centering
\includegraphics[scale=0.32]{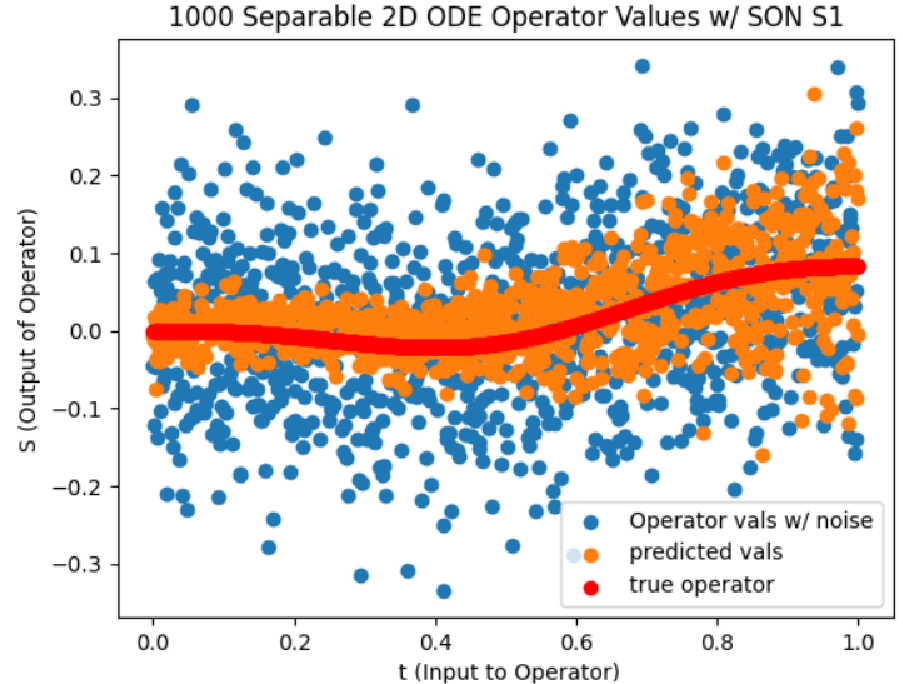}
\end{minipage}%
\begin{minipage}{0.48\textwidth}
     \centering
\includegraphics[scale=0.32]{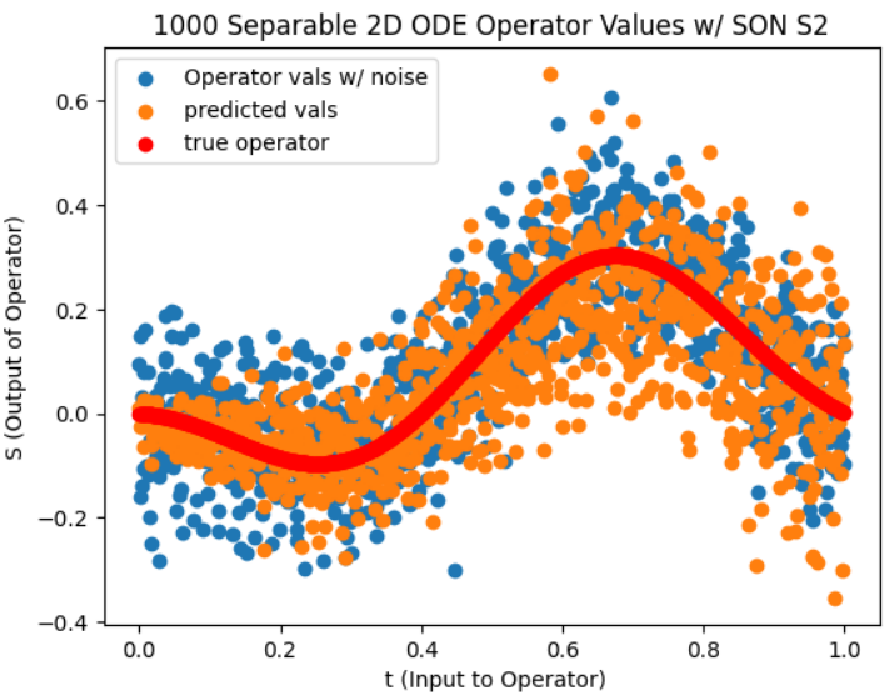}
\end{minipage}

\begin{minipage}{0.48\textwidth}
     \centering
\includegraphics[scale=0.32]{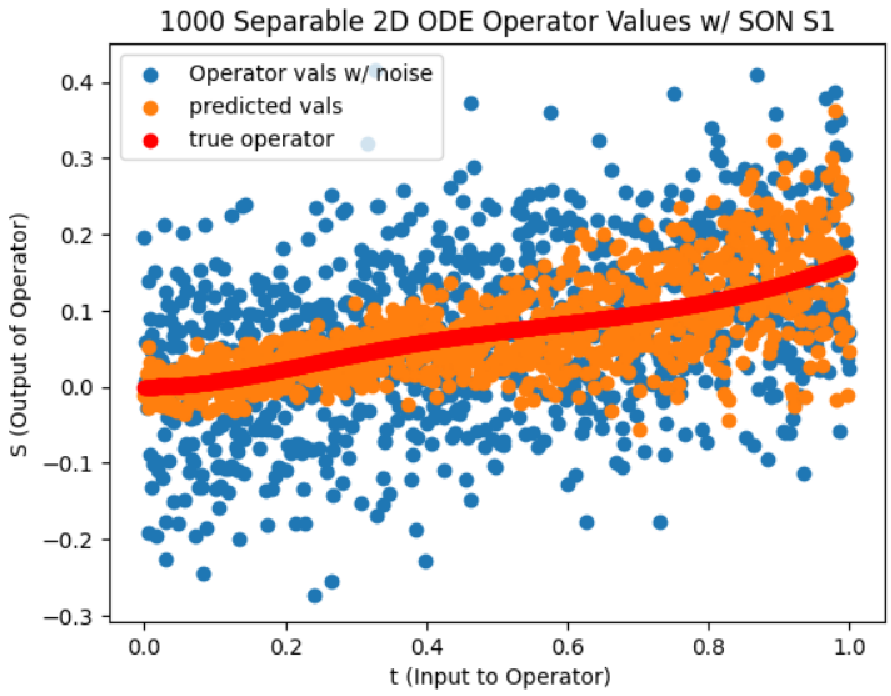}
\end{minipage}%
 \begin{minipage}{0.48\textwidth}
     \centering
\includegraphics[scale=0.32]{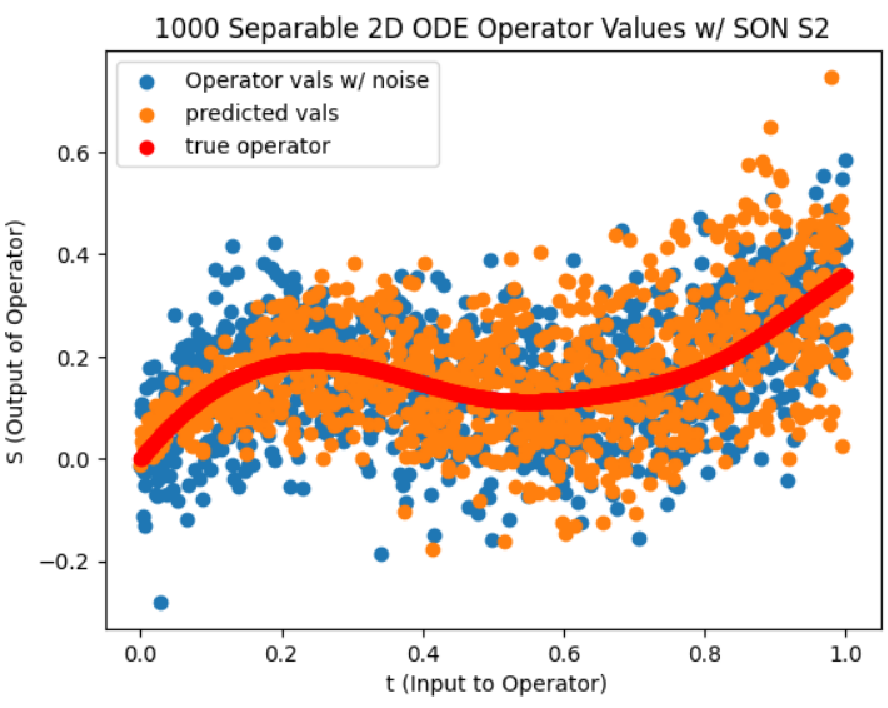}
\end{minipage}

\begin{minipage}{0.48\textwidth}
    \centering
\includegraphics[scale=0.32]{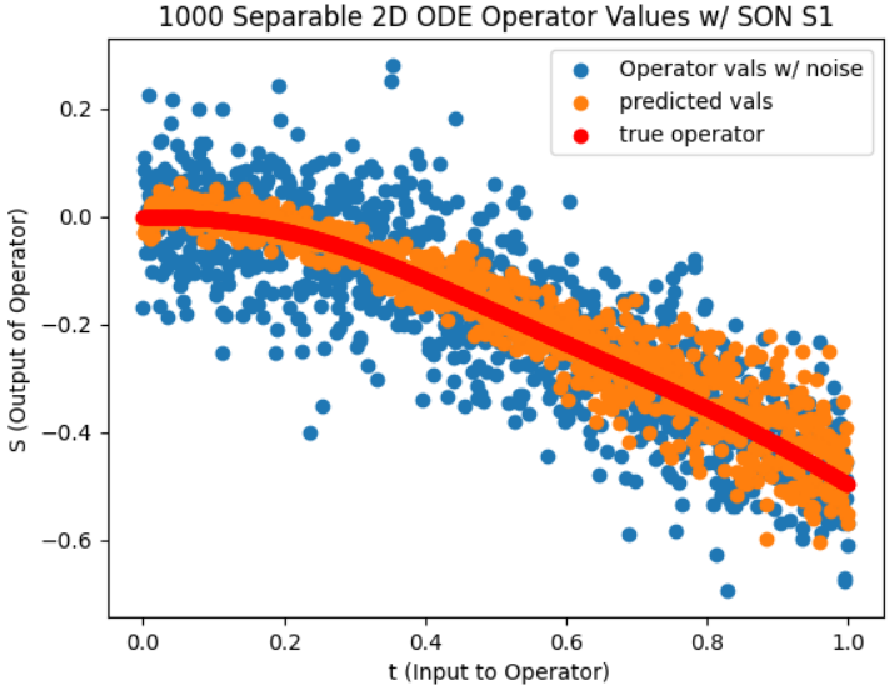}
\end{minipage}%
\begin{minipage}{0.48\textwidth}
\centering
\includegraphics[scale=0.32]{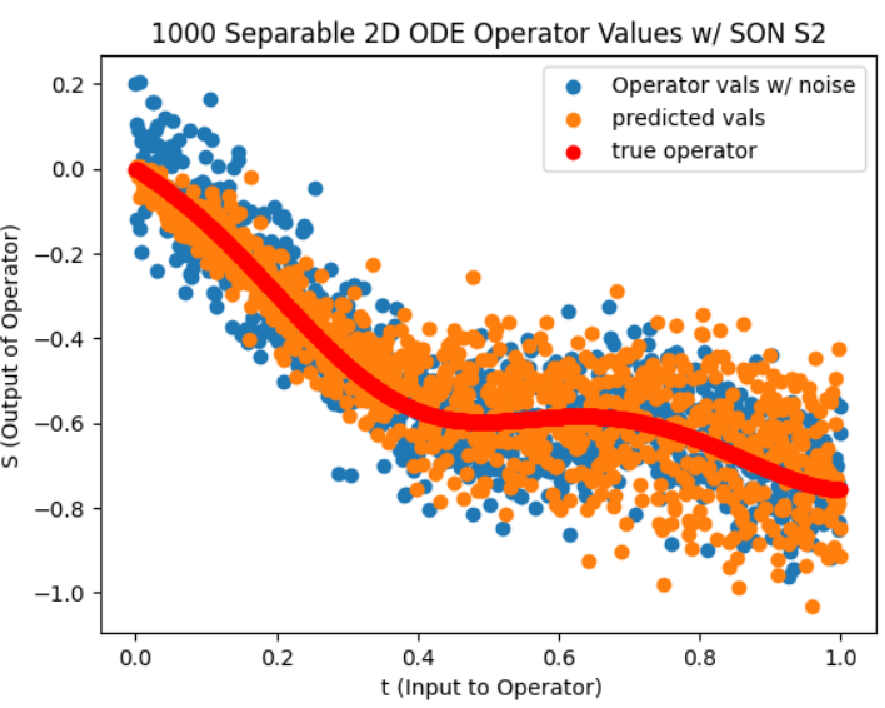}
\end{minipage}
        \caption{\small Pairs of SON predicted 2D ODE trajectories with 3 random test $u(x)$.}
        \label{fig:2dODEplots}
        \vspace{-0.5cm}
\end{figure}
SON achieved a final MSE of $0.0162$ after 2000 epochs, as seen in Figure \ref{fig:2dodeloss}. Figure \ref{fig:2dODEplots} shows the pairs of predicted trajectories for three randomly sampled $u(x)$ in orange, as well the true trajectories (red) and with noise (blue). It is clear that SON is able to capture the true trajectory within the orange prediction band, as well as quantify the majority of noise on both dimensions. Additionally, after performing the same standard deviation test, SON recovered 0.069 and 0.17 as the noise level estimations for dimensions 1 and 2, which are both closely surrounding the correct scaling factor of $\alpha =0.1$. This averages to an estimated noise quantification  estimate of 0.12 averaged on both dimensions, or the noise scaling in the 2D spatial domain.


\subsection{Double Integral with Noise}\label{DoubleInt}
\begin{figure}[h!]
\begin{subfigure}{0.5\textwidth}
    \includegraphics[scale=0.45]{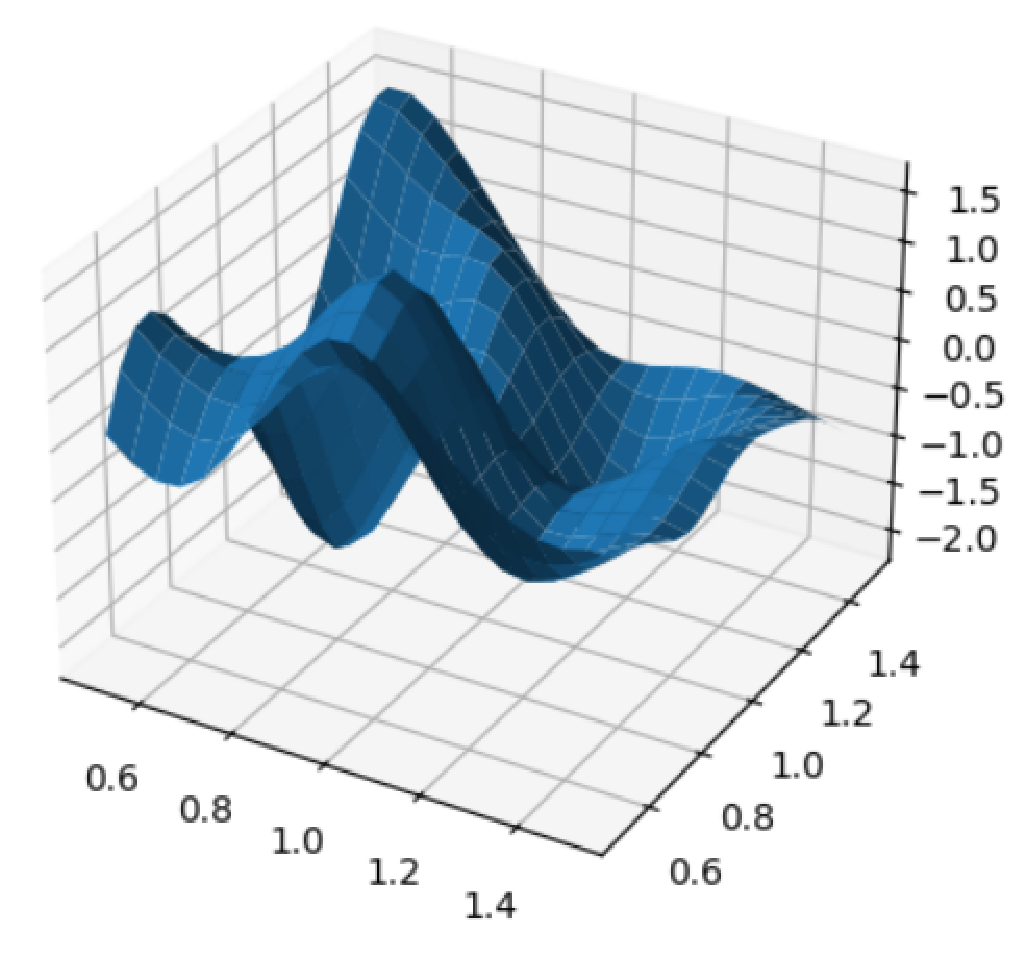}
    \caption{}
    \label{fig:3d U ex}
\end{subfigure}
\begin{subfigure}{0.5\textwidth}
    \includegraphics[scale=0.37]{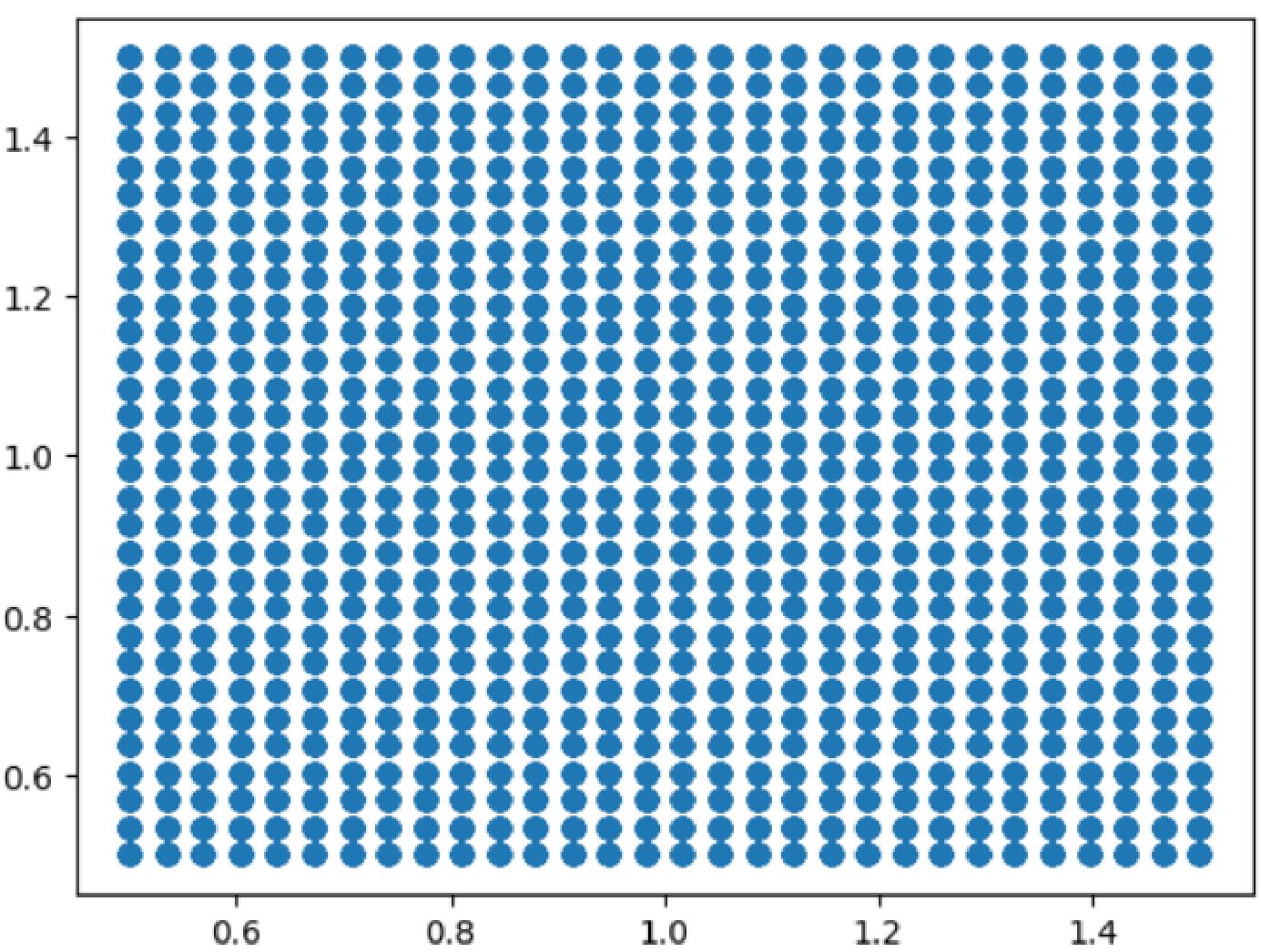}
    \caption{}
    \label{fig:op domain 3d}
\end{subfigure}
\hfill
    \caption{(\ref{fig:3d U ex}) Example of $u(x)$ from training set; (\ref{fig:op domain 3d}) Operator output domain.}
    \vspace{-0.3cm}
\end{figure}

We next consider the following double integral with noise:
$$s: u(\pmb{x}) \to G(u)(\pmb{y}) = \int_0^{y_1}\int_0^{y_2} u(\tau_1, \tau_2)d\tau_1 d\tau_2 + \alpha\epsilon,$$
where $\pmb{x} \in \mathbb{R}^2$, $\pmb{y} = (y_1, y_2) \in \mathbb{R}^2$ and $\epsilon \sim N(0,1)$ noised scaled by $\alpha = 0.05$. Our map is from a space of 3D input $u(\pmb{x})$ functions to a space of 3D operator output functions $s$. 
$u(\pmb{x})$ is once again sampled from a mean-zero GRF with the RBF (radial basis function) kernel with $l=0.2$. However, $[0.5, 1.5]$ was now partitioned into 20 sensors along each dimension, resulting in 400 total sensors for each $u(\pmb{x})$. The training and test sets had 100 and 20 $u(\pmb{x})$ each respectively, which is close to the $80\%/20\%$ train/test split commonly used. The first $u(\pmb{x})$ example in the training set is shown in Figure \ref{fig:3d U ex}.

The operator output domain is now $\pmb{y} \in [0.5, 1.5] \times [0.5, 1.5]$ to avoid an infinite values or discontinuities near 0. This domain was partitioned to give 30 input values along each dimension, resulting in $30 \times 30 =900$ $\pmb{y}$ for both test and training (Figure \ref{fig:op domain 3d}). 

\begin{figure}[h!]
\centering
\begin{minipage}{0.33\textwidth}
    \includegraphics[scale=0.35]{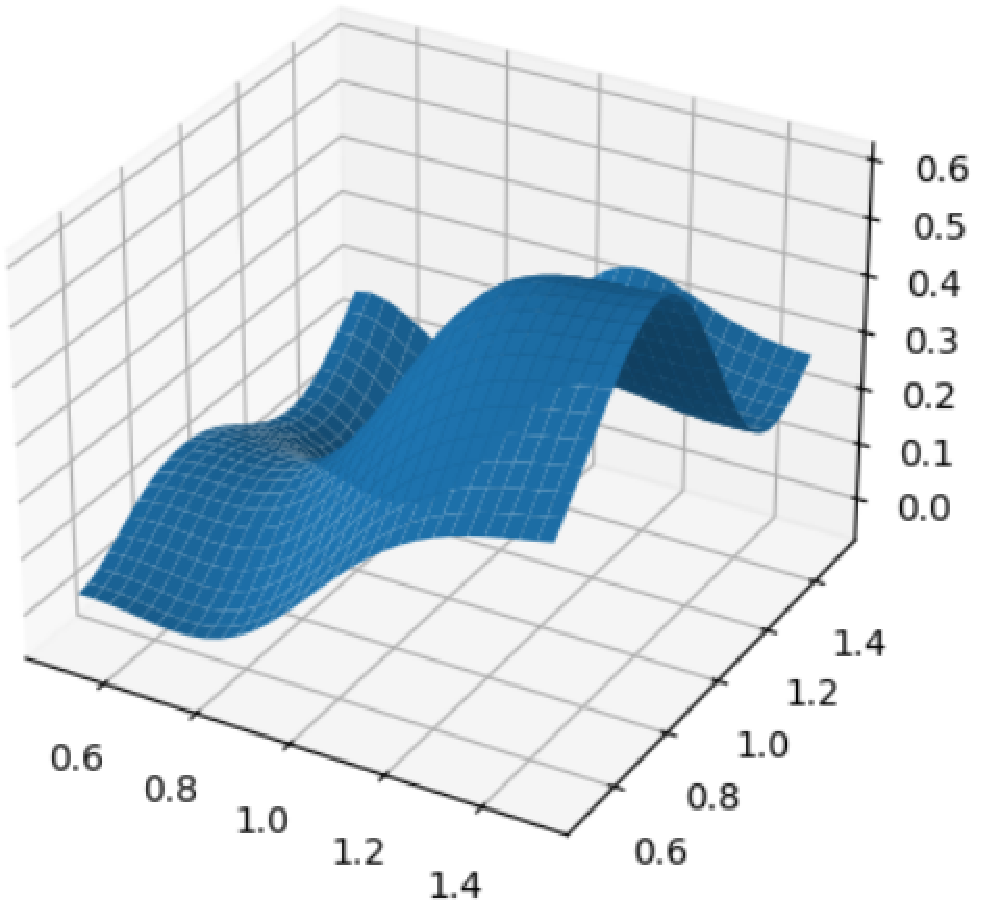}
    \label{fig:S 3d ex}
\end{minipage}%
\begin{minipage}{0.33\textwidth}
    \includegraphics[scale=0.35]{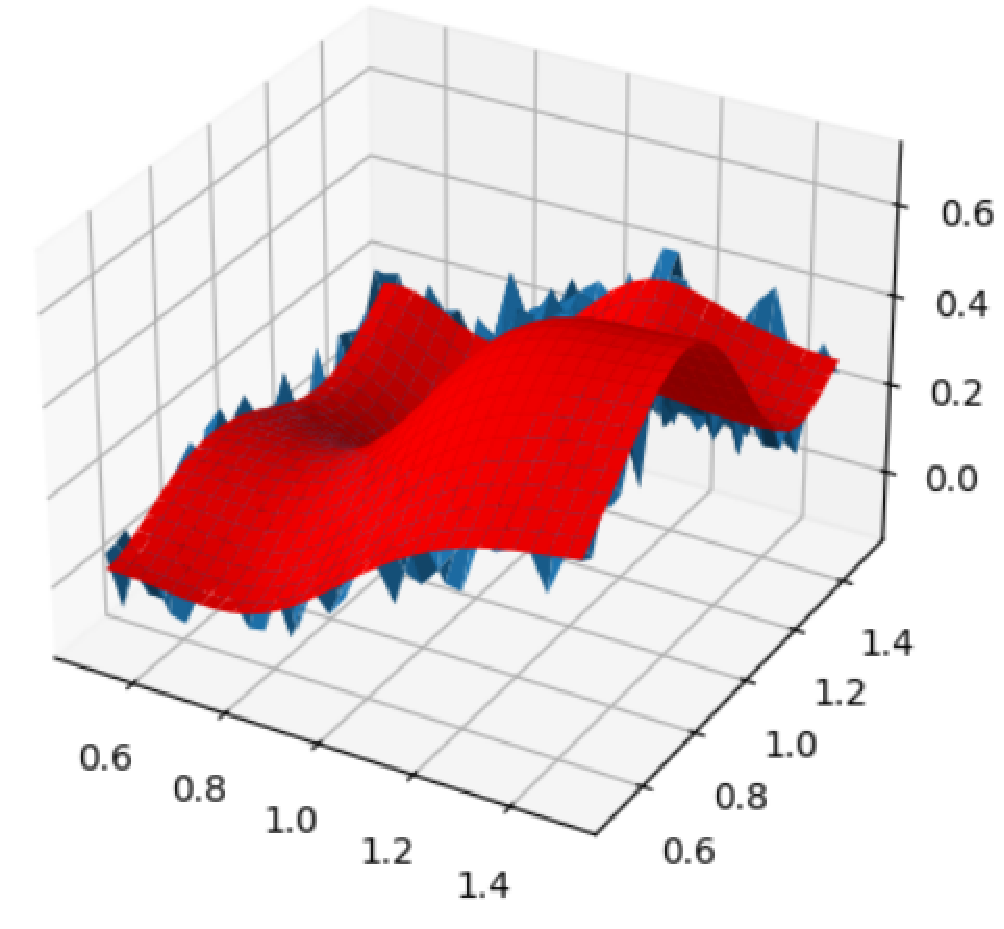}  
    \label{fig:S 3d noise ex}
\end{minipage}%
\begin{minipage}{0.33\textwidth}
 \hspace{0.3cm}\includegraphics[scale=0.35]{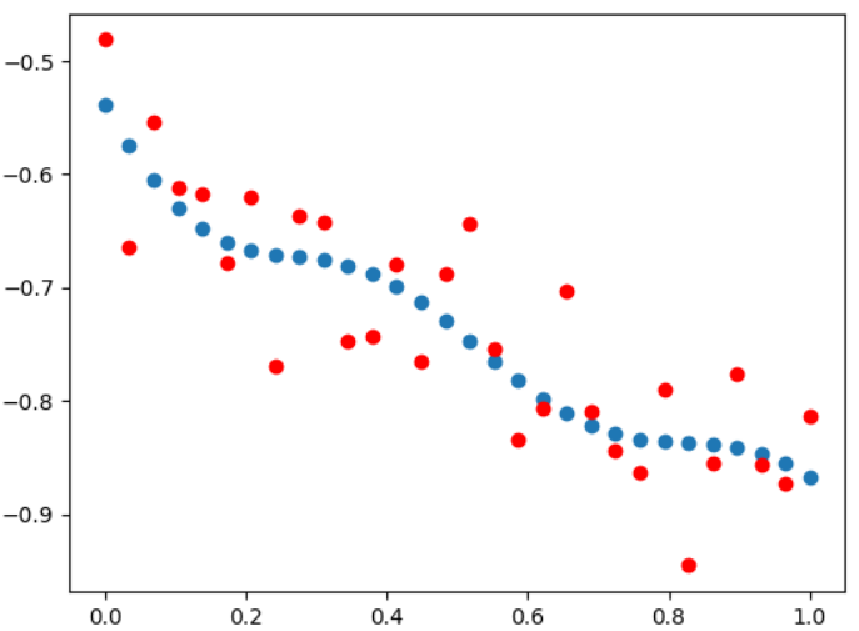}
    \label{fig:S 3d noise ex cross section}
\end{minipage}
    \caption{\small (Top Left) Random output surface of double integral operator; (Top Right) Double integral with scaled $N(0,1)$ noise (blue), true operator output(red); (Bottom) Cross section of operator output surface along $t_1=0.5$ axis.}
    \label{DoubleInt_Noisy}
    \vspace{-0.4cm}
\end{figure}

The cartesian product of the $u(\pmb{y})$ and $\pmb{y}$ was then taken and the double integral of each $(\pmb{y}, u(\pmb{x}))$ was computed numerically, resulting operator output surfaces as shown in the top left panel of Figure~\ref{DoubleInt_Noisy}. Noise was then added as described above (the top right panel of Figure~\ref{DoubleInt_Noisy}, where red is the true operator output and blue the noisy output used during SON training and testing).

\begin{figure}[h!]
    \centering
    \includegraphics[scale =0.35]{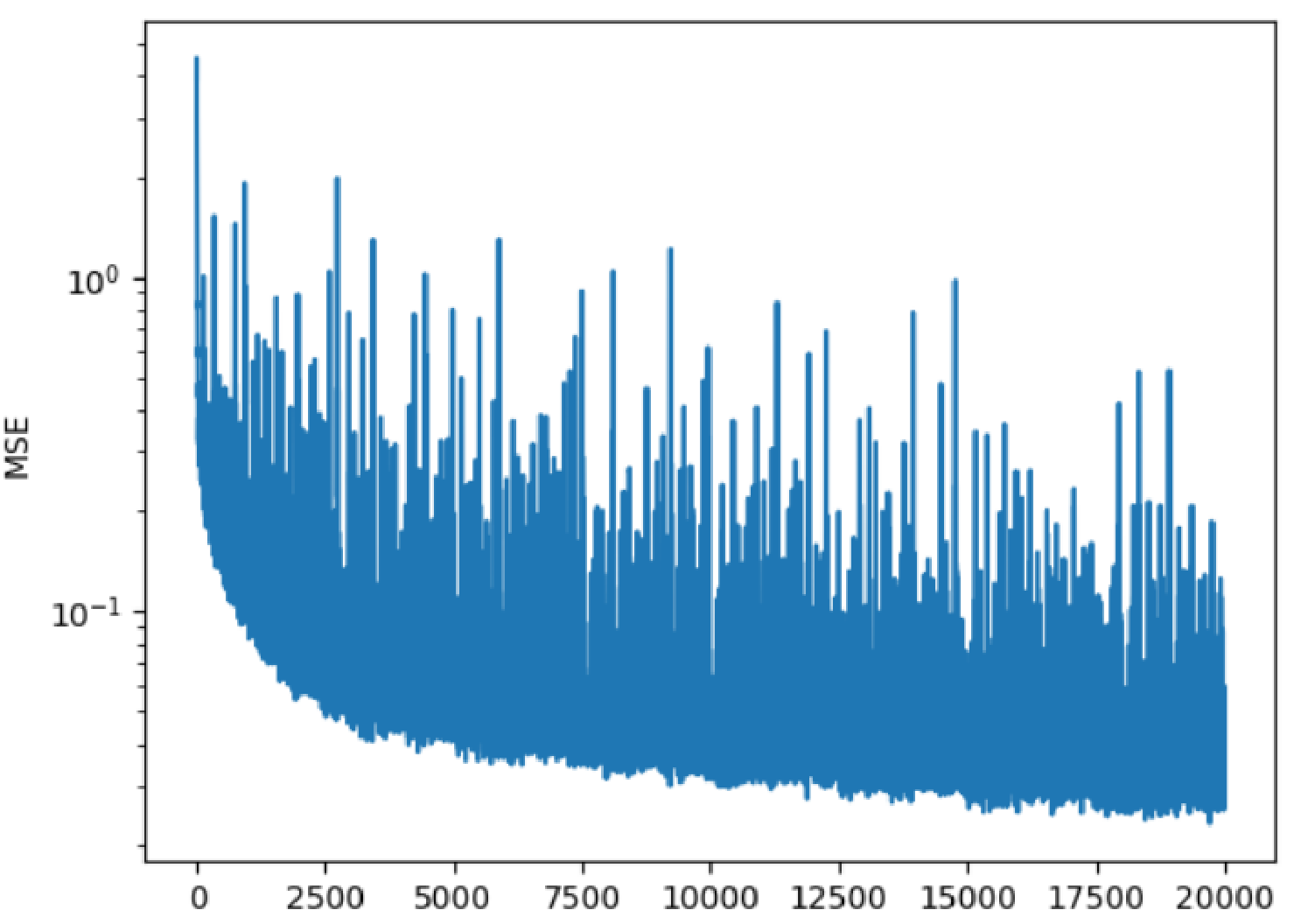}
    \caption{\small Training MSE loss for double integral operator with noise over 200 epochs and 100 batches per epoch.}
    \label{fig:lossdoubleint}
    \vspace{-0.4cm}
\end{figure}
As before, the trunk net was a 2 layer dense architecture, now with input dimension two instead of one and sigmoid activation. However, due to the size of the input data to the SNN and its structure, the both the drift $\mu$ and diffusion $\sigma$ were changed to stacks of convolutional layers. A single input to the SNN was hence $20 \times 20$ points from surface $u(x)$. For $\mu$, a single layer with a kernel window size of 3 was employed with ReLU activation, while for $\sigma$, a single layer with kernel window size 3 and arctan activation was used, followed by a dropout layer with $p=0.9$. This helped to prevent the predicted noise level from exploding after many training epochs. This gives the following form for the SNN forward discretized SDE:
 $$u_{n+1} = u_n+ \mu(u_n, \theta)\Delta n + \sigma(u_n, \theta)\sqrt{\Delta n}\epsilon, \; n=0,\hdots, N-1,$$
where $\theta$ are trainable weights. Five layers were used for the SNN, resulting in each pseudo-time-step $\Delta n=\frac{1}{5}$.

Additionally, two projection layers were used in the SNN prior to the SDE propagation, where max pooling was utilized to further simplify the input function $u(x)$ and assist with surface feature extraction. The forward SDE output was then max-pooled a final time before being flatten for combination with the trunk net output.

SON was trained with ADAM for 200 epochs, with learning rate 0.001 reduced to 0.9 of the previous level by the scheduler every 25 epochs. The batch size was reduced to 900, meaning the double integrals of a single $u(x)$ were fed in per batch. 

Final MSE for the training set was 0.0296 (Figure \ref{fig:lossdoubleint}), with 0.043 MSE on the testing set. This indicates that SON is generalizing well to unknown data, and final MSEs are within the scaling factor of 0.05 of the added noise.  However, although Figure \ref{fig:lossdoubleint} does indicate convergence of the loss, it still fluctuates wildly despite utilization of scheduler to periodically reduce the learning rate. This is a common problem in operator learning, as even the initial DeepONet experiments in~\cite{VAnh2024, Lu2019}. 

\begin{figure}[h!]
\centering
   \begin{minipage}{\textwidth}  \includegraphics[scale=0.3]{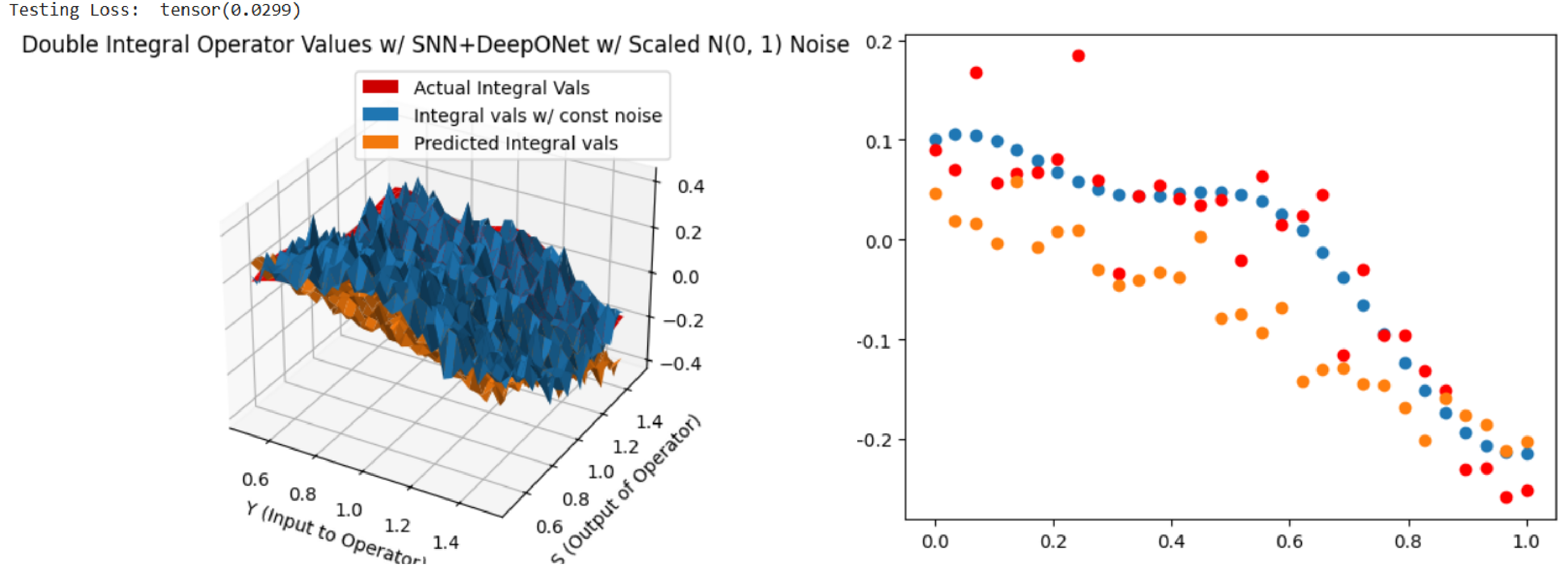}
    \label{fig:pred 4 pair}
\end{minipage}
\begin{minipage}{\textwidth}
\includegraphics[scale=0.3]{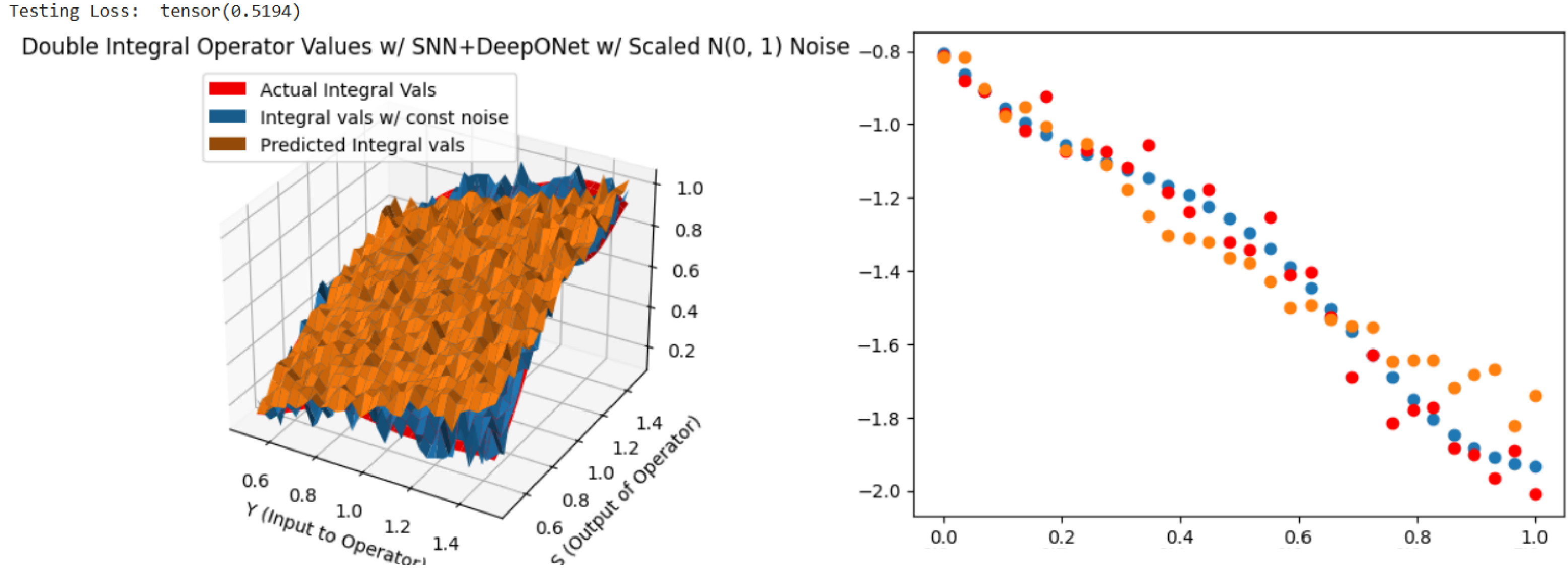}
    \label{fig:pred 9 pair}
\end{minipage}
\begin{minipage}{\textwidth} \includegraphics[scale=0.3]{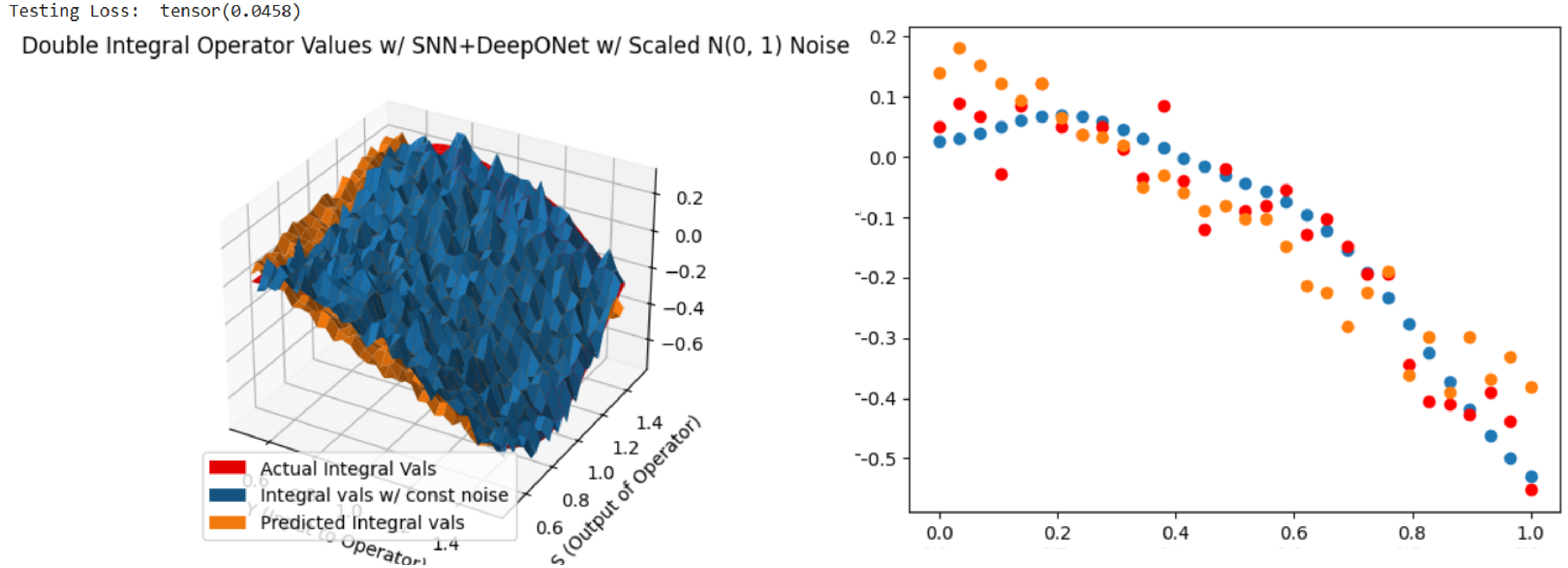}
    \label{fig:pred 10 pair}
\end{minipage}
    \caption{\small Three random SON double integral operator with noise test set predictions compared with true operator output (left) and corresponding random cross sections (right).}
    \label{fig:doubleintplots}
\end{figure}

As Figure \ref{fig:doubleintplots} shows for three random test set examples, SON is able to capture the mean underlying surface produced by the output of the double integral operator as well as quantify the uncertainty present.

Additionally, the standard deviation at all points of all operator outputs for all input functions $u(x)$ was calculated 20 times and the results averaged. This gave an estimate of 0.0359 on the noise level present, when then $N(0,1)$ noise was scaled by 0.05 before being added to the operator output. Given that these values are relatively close, this indicates that SON is able to correctly quantify the uncertainty present in this operator, confirming the intuition generated by the plots.



\subsection{Stochastic Elliptic Equation with Multiplicative Noise}
\begin{figure}[h!]
\begin{minipage}{0.48\textwidth}
\includegraphics[scale=0.28]{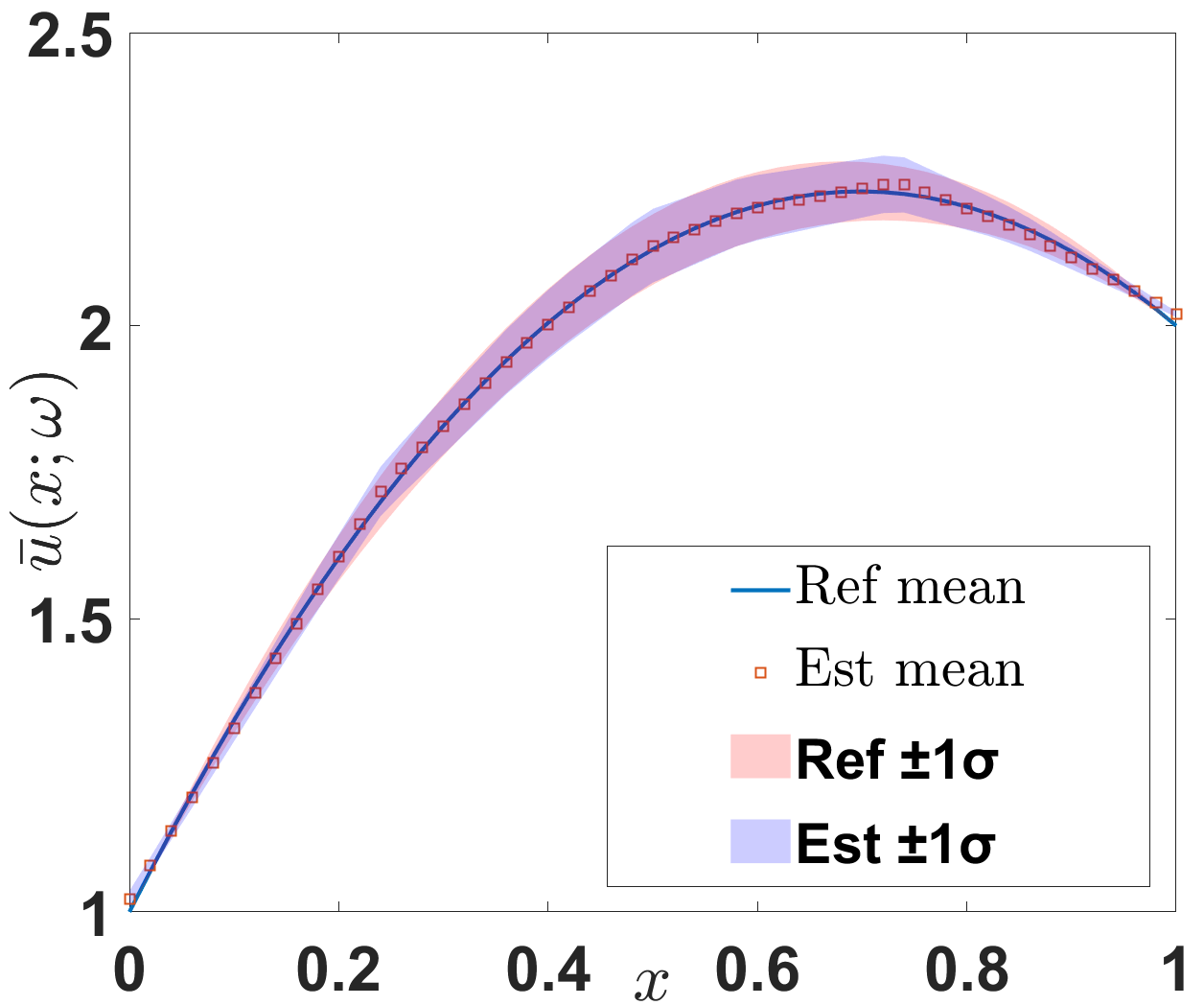}
\end{minipage}%
\begin{minipage}{0.48\textwidth}
\includegraphics[scale=0.28]{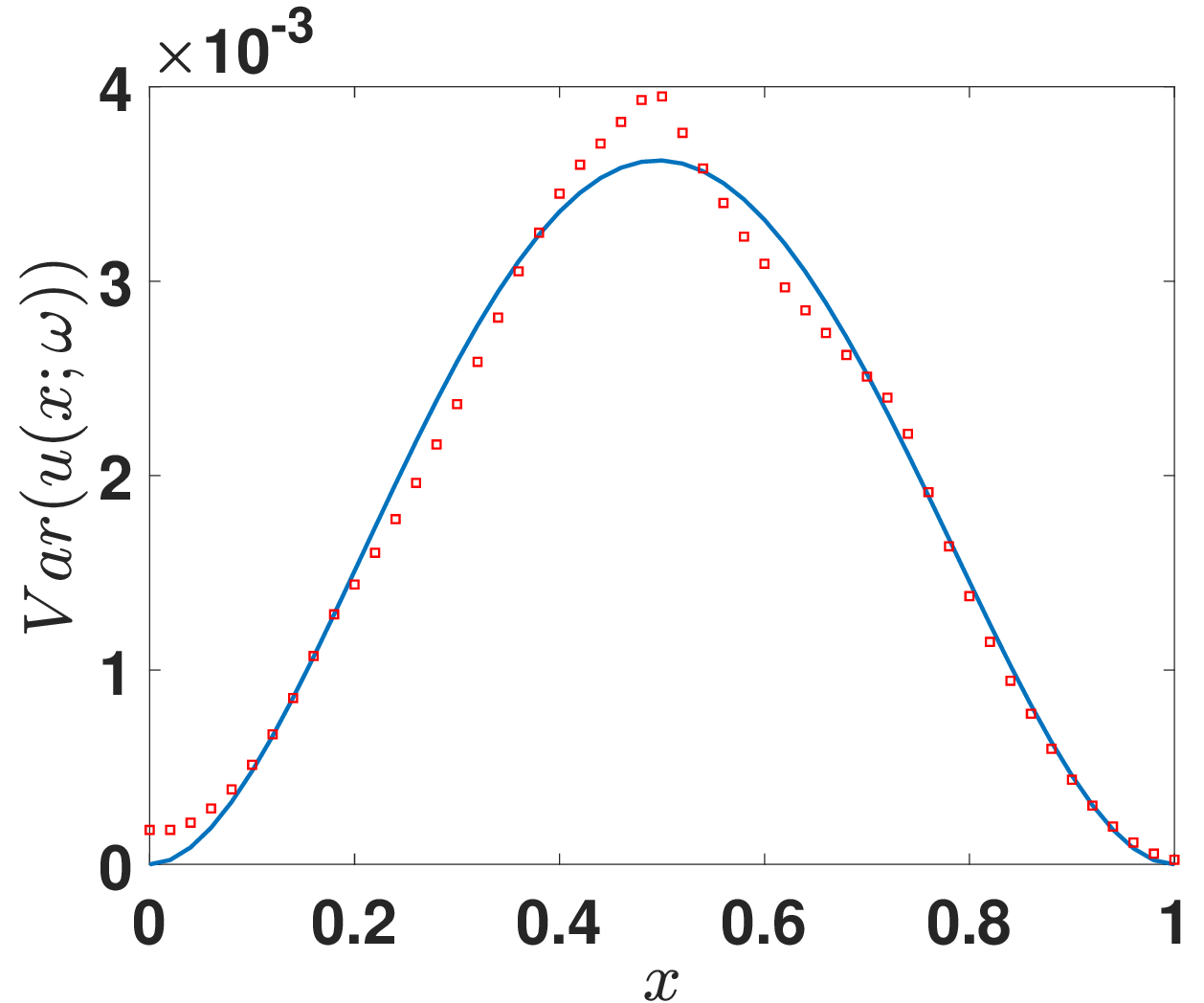}
\end{minipage}
\caption{\small With $l=1.5$ (Left) Mean and confidence band. (Right) Spread of the data.}
\label{Sec2_MeanVar}
\vspace{-0.3cm}
\end{figure}

\begin{figure}[h!]
\begin{minipage}{0.33\textwidth}
\includegraphics[scale=0.25]{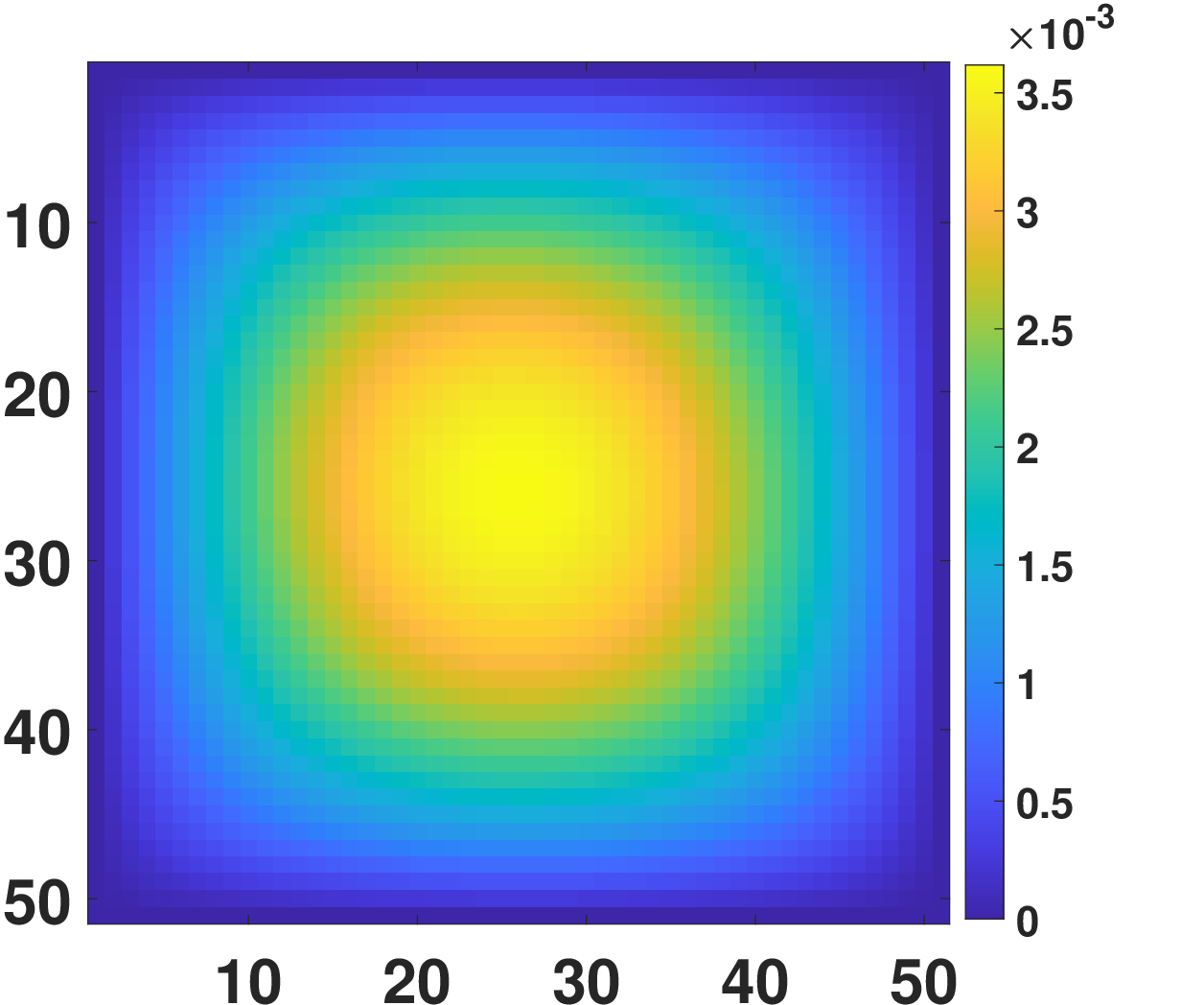}
\end{minipage}%
\begin{minipage}{0.33\textwidth}
\includegraphics[scale=0.25]{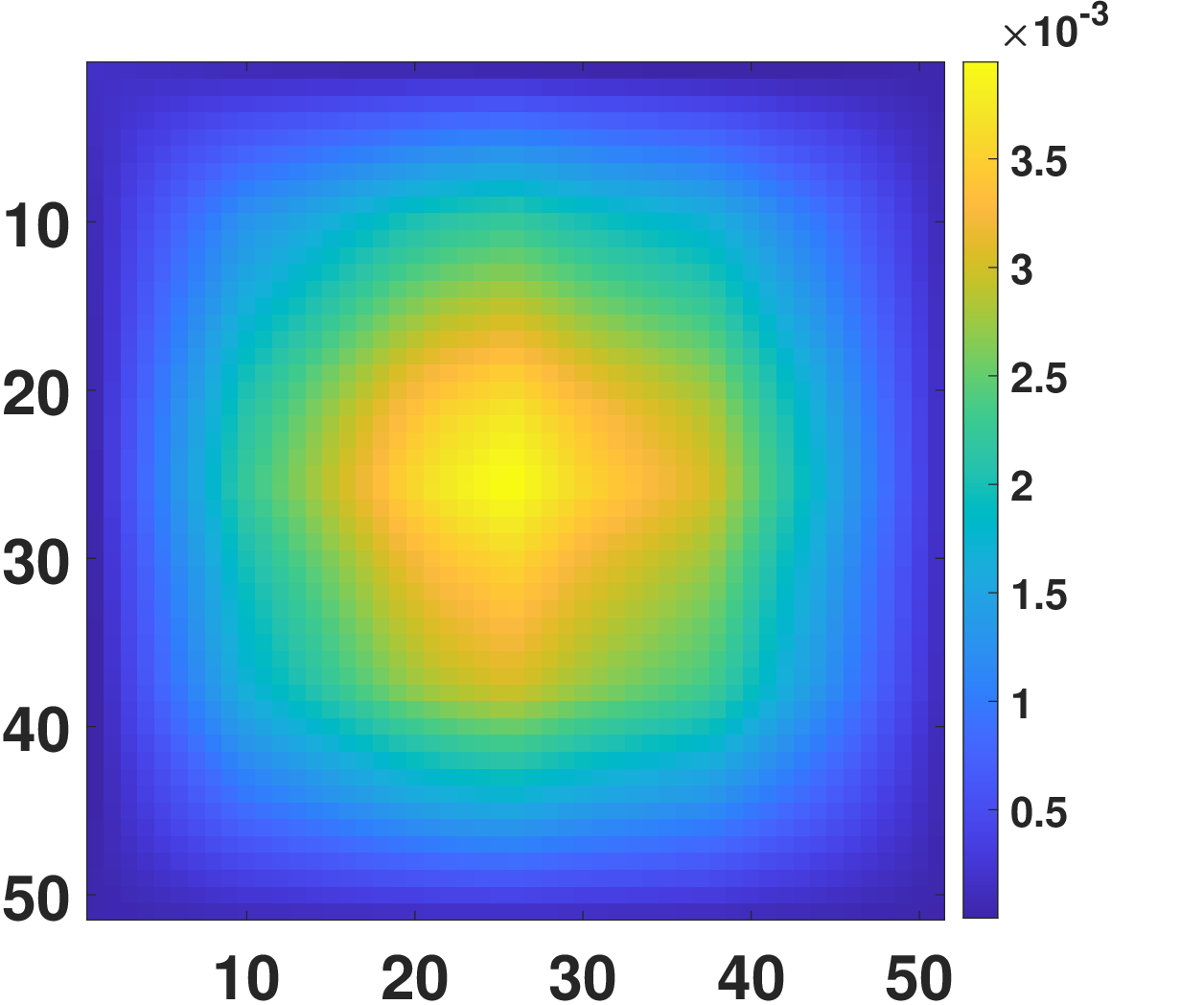}
\end{minipage}%
\begin{minipage}{0.33\textwidth}
\includegraphics[scale=0.25]{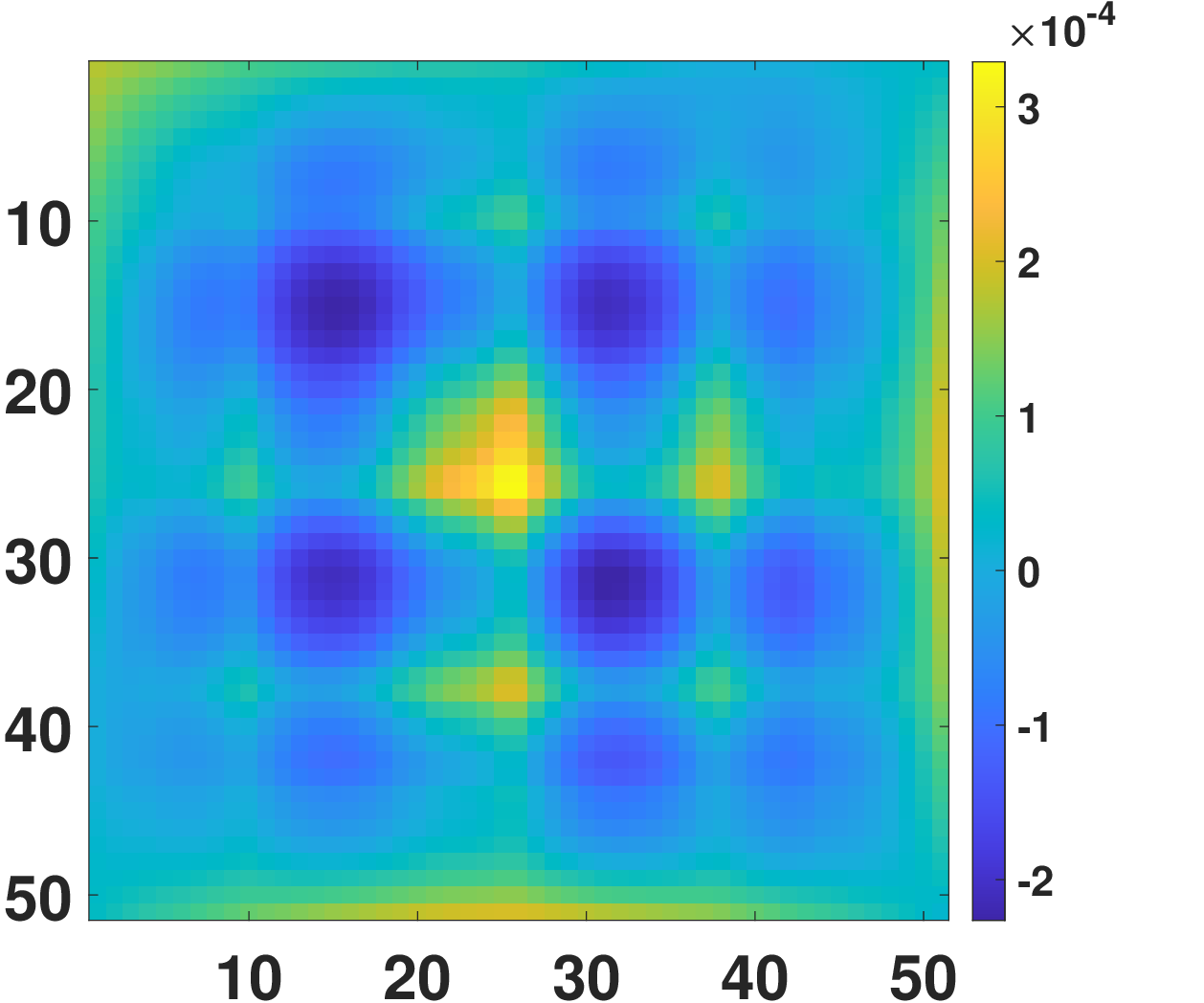}
\end{minipage}
\caption{\small Covariance with $l=1.5$: (First) Reference. (Second) Estimation. (Third) Difference.}
\label{Sec2_Covariance}
\end{figure}

We adopt a test case from~\cite{Lu2021} and consider  the solution operator of the following one-dimensional SPDE:
\begin{equation}
\label{StoElliptic}
\begin{array}{l}
\text{div}\left(e^{b(x; \omega)} \nabla u(x; \omega)\right) = f(x), \; x \in (0, 1] \; \text{and} \; \omega \in \Omega,
\end{array}
\end{equation}
\begin{figure}[h!]
\begin{minipage}{0.47\textwidth}
\includegraphics[scale=0.26]{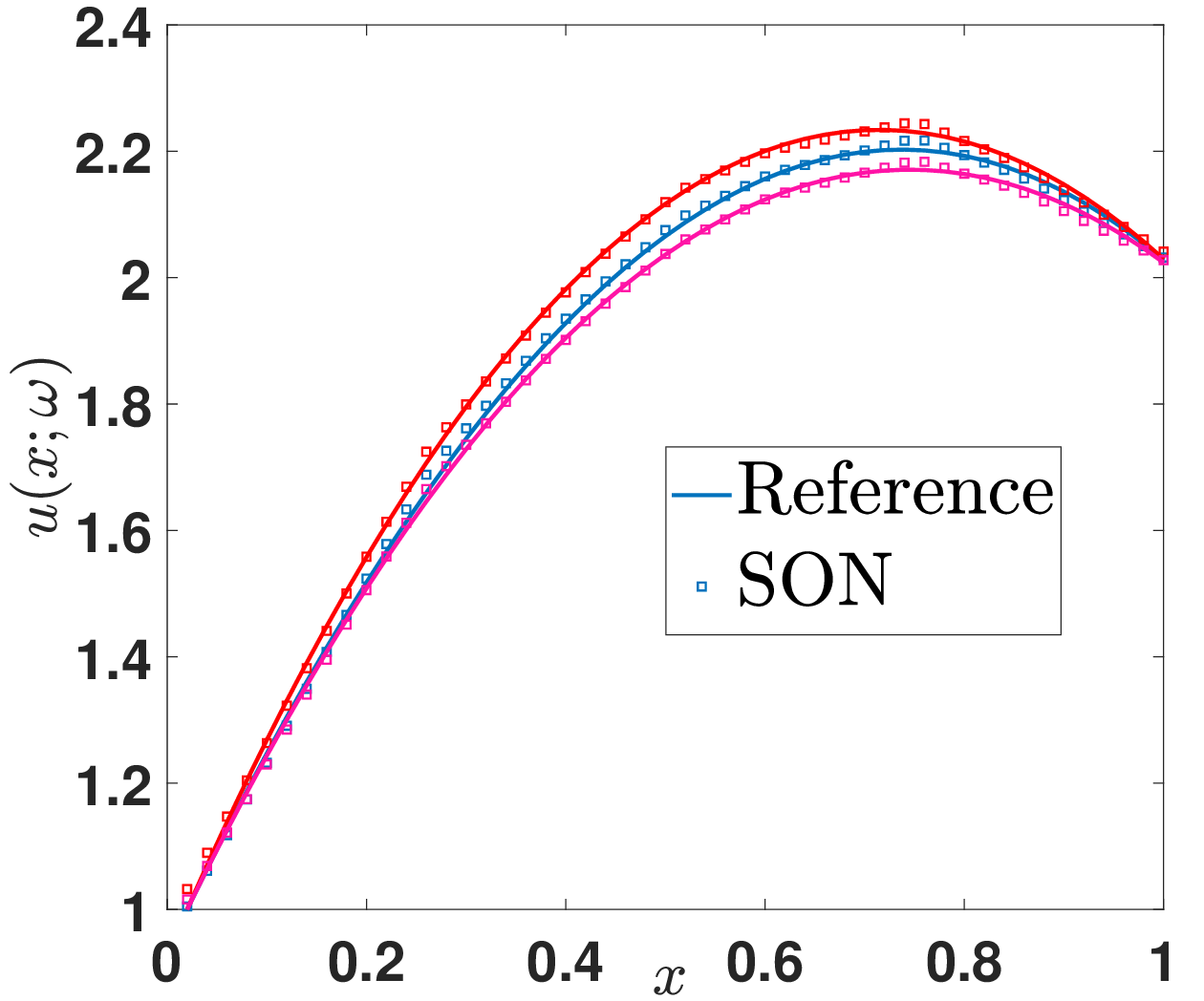}
\end{minipage}%
\begin{minipage}{0.47\textwidth}
\includegraphics[scale=0.27]{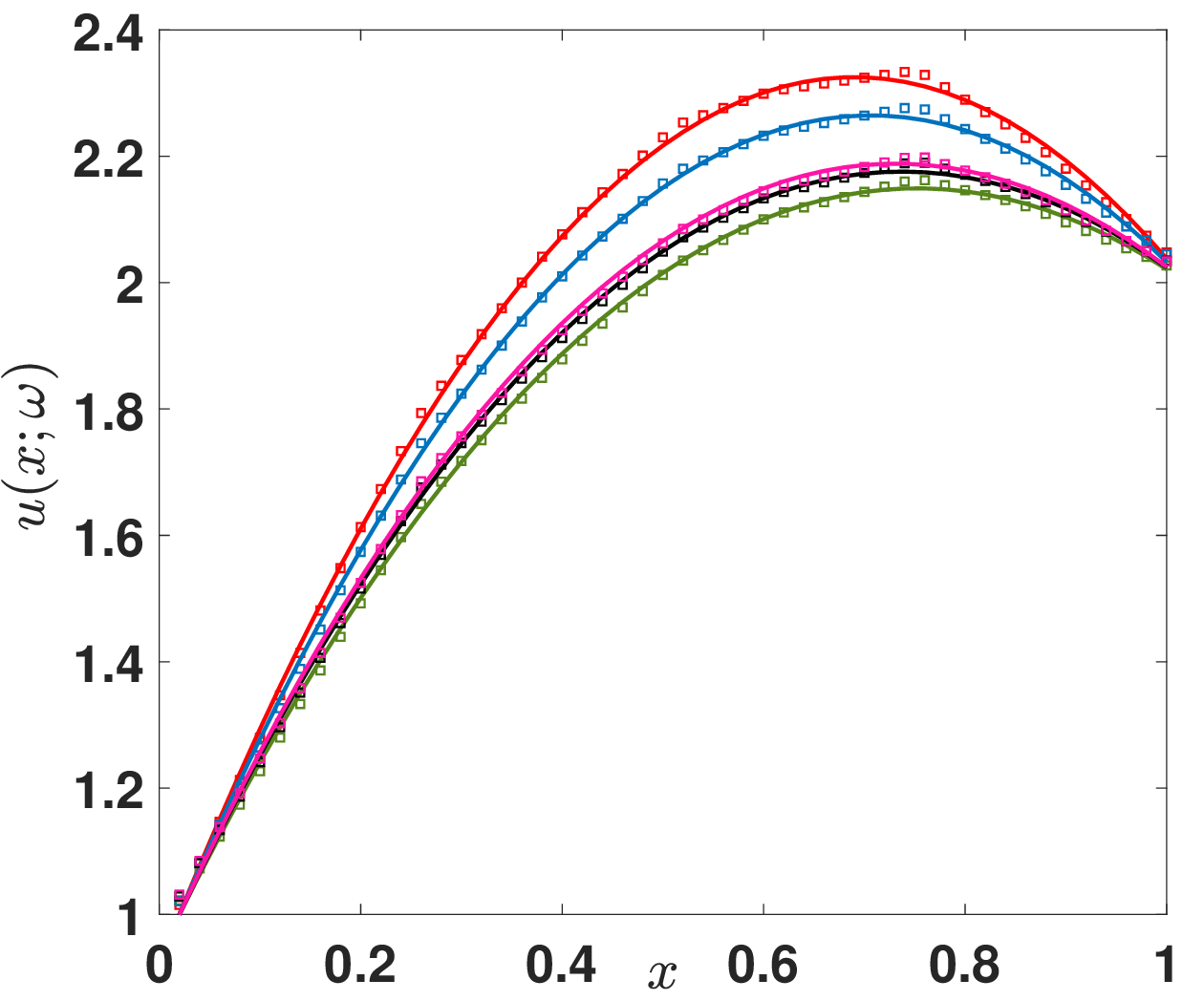}
\end{minipage}%

\begin{minipage}{0.47\textwidth}
\includegraphics[scale=0.26]{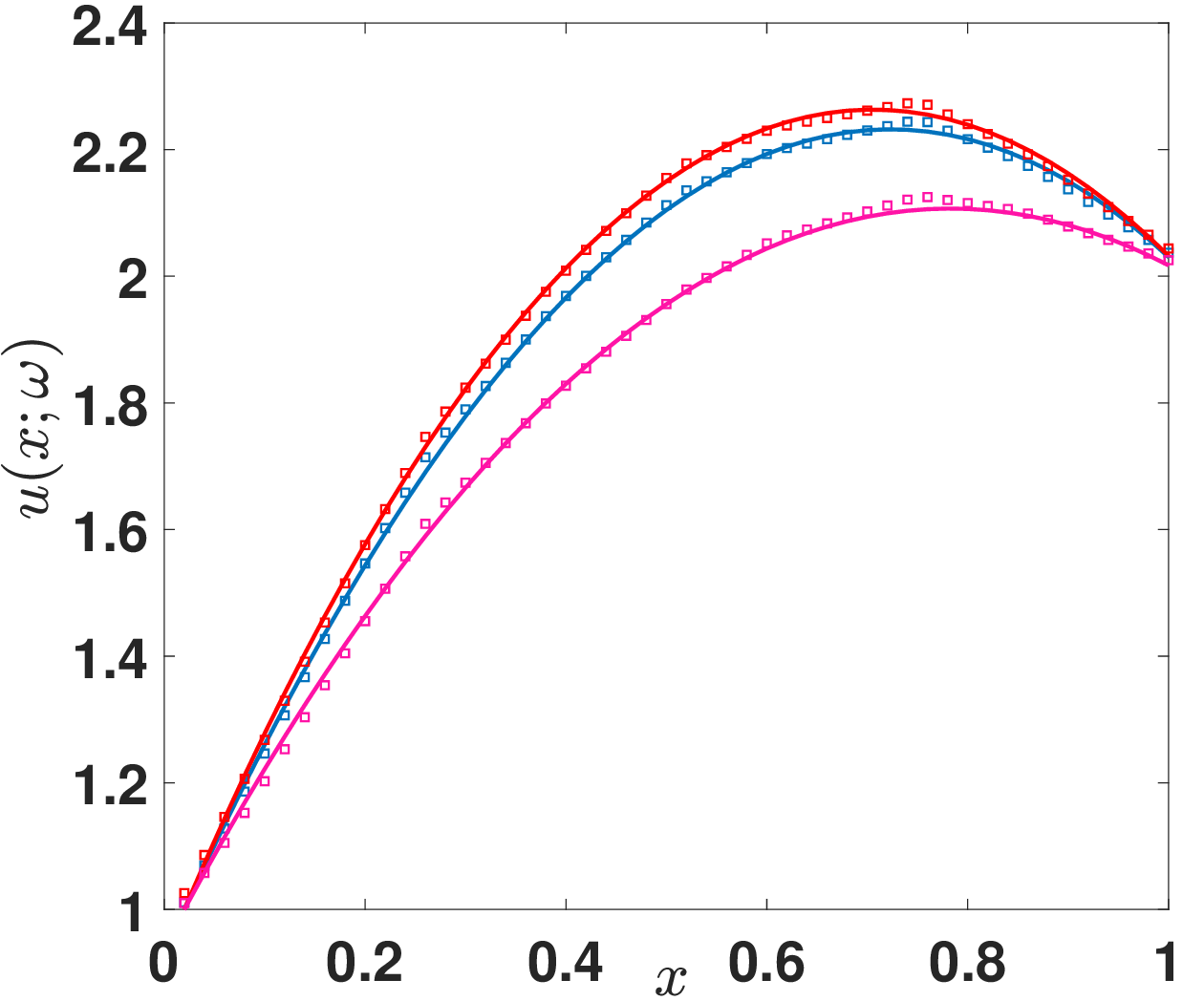}
\end{minipage}%
\begin{minipage}{0.47\textwidth}
\includegraphics[scale=0.27]{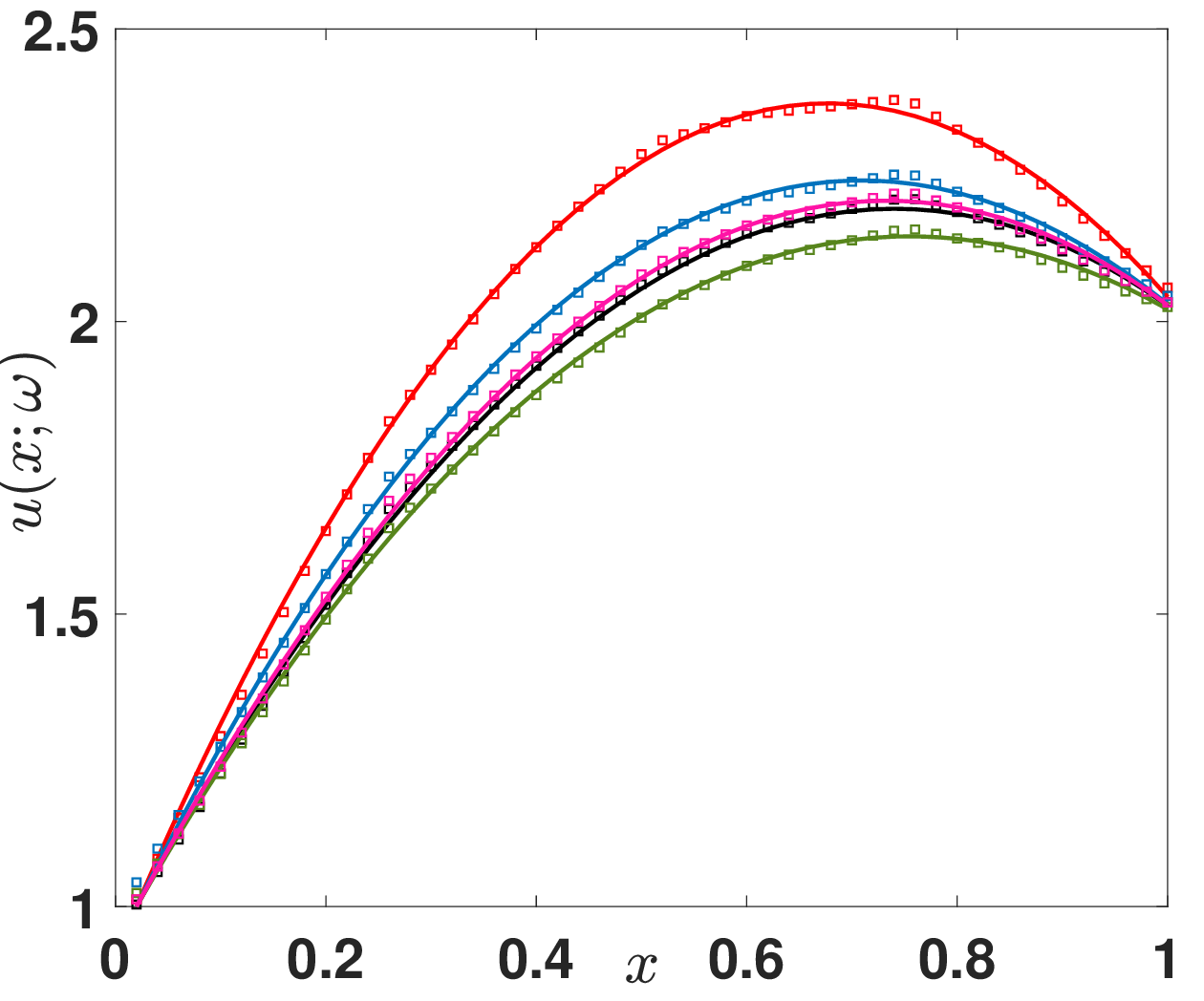}
\end{minipage}
\caption{\small The prediction
for 8 different random samples from $b(x; \omega)$ with (Top) $l=1.5$. (Bottom) $l=1.7$.}
\label{Sec2}
\vspace{-0.3cm}
\end{figure}
with Dirichlet boundary conditions $u(0) =0, \; u(1) =1$, $f(x) =5$, and $$b(x; \omega) \sim \mathcal{G}\mathcal{P}\left(b_0(x), \; \text{Cov}(x_1, x_2)\right),$$
where the mean $b_0(x) =0$ and the covariance function is 
\begin{align*}
\begin{array}{l}
\text{Cov}(t_1, t_2) = \sigma^2 \text{exp}\left(-\dfrac{\vert\vert t_1-t_2\vert\vert^2}{2l^2}\right),
\end{array}
\end{align*}
with $\sigma=0.1$ and $l \in [1, 2]$. Unlike~\cite{Lu2021}, we do not have to use Karhunen–Lo\`{e}ve (KL) expansion for $k(t; \omega)$ to inject randomness into the trunk network. Instead, stochasticity is introduced automatically by propagating $k(t; \omega)$ through the SNN layers of the branch network. As a result, the input for the branch network is the random process $k(t\; \omega)$, while the trunk network only needs to take $t$ as input. This significantly reduces the dimension of the training process. We train the SON with 5000 different $k(t; \omega)$ with $l$ randomly sampled in $[1, 2]$ and for each $k(t; \omega)$, we only use one realization. The prediction for 8 different random samples from $k(t; \omega)$ with $l=1.5$ and $l=1.7$ are shown in Figure~\ref{Sec2}. To further quantify the uncertainty, we generate $1000$ samples and generate the prediction band in Figure~\ref{Sec2_MeanVar} to visualize the spread of the data. The mean and variance of these samples are also shown in Figure~\ref{Sec2_MeanVar}. Finally, the corresponding covariances and their differences are shown in Figure~\ref{Sec2_Covariance}.

It can be observed from the above figures that the estimated solution operator can reflect well the spread of the solution ensemble, as evidenced by its ability to match the sample variance and covariance of the reference data.

\subsection{Comparison with Vanilla DeepONet}\label{compDeepONet}
In addition to demonstrating the effectiveness of SON when learning noisy operators, we also compare our method's performance against a vanilla DeepONet. The DeepONet will compare its predictions against the noisy operator output $G(u)(y)$ during training, just like SON. Since vanilla DeepONet has no mechanism to probabilistically learn the output noise, all predictions will be fully deterministic. 

\begin{figure}[h!]
    \centering
    \begin{minipage}{0.42\textwidth}
        \hspace{-0.5cm}\includegraphics[scale=0.37]{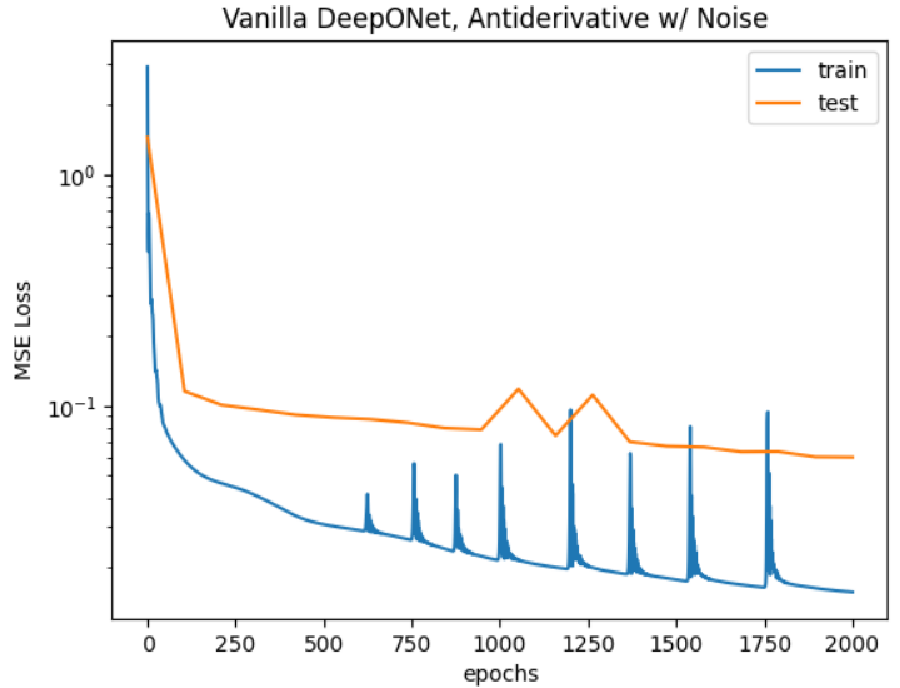}
    \end{minipage}%
    \begin{minipage}{0.42\textwidth}
        \includegraphics[scale=0.37]{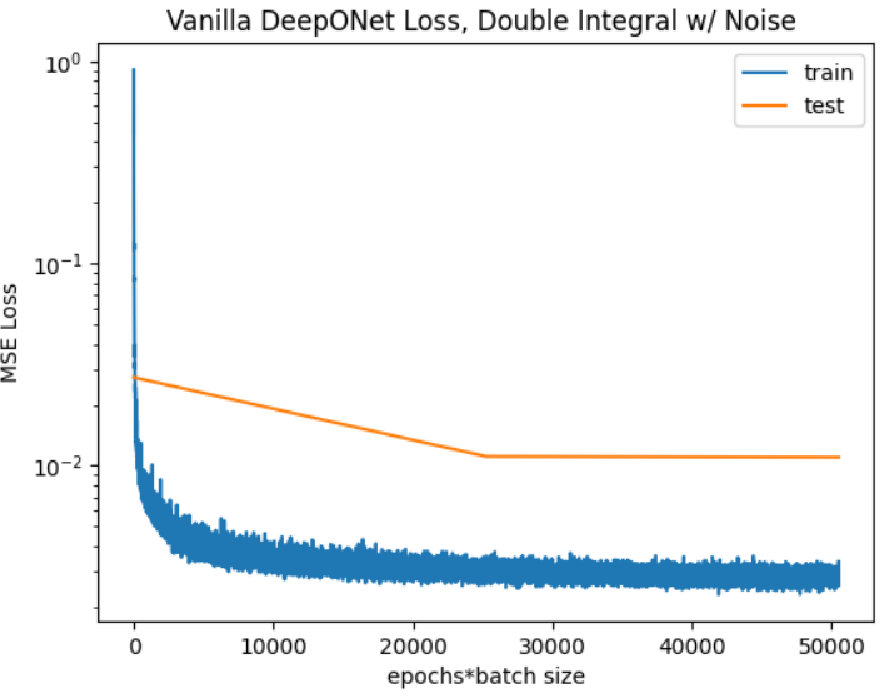}
    \end{minipage}
    \caption{Final Antiderivative Operator MSE Losses: 0.016 Training, 0.06 Testing; Double Integral: 0.002 Training, 0.005 Testing.}
    \label{fig:VdeeponetANtiloss}
\end{figure}

\begin{figure}[h!]
    \centering
    \begin{minipage}{0.45\textwidth}
\includegraphics[scale=0.38]{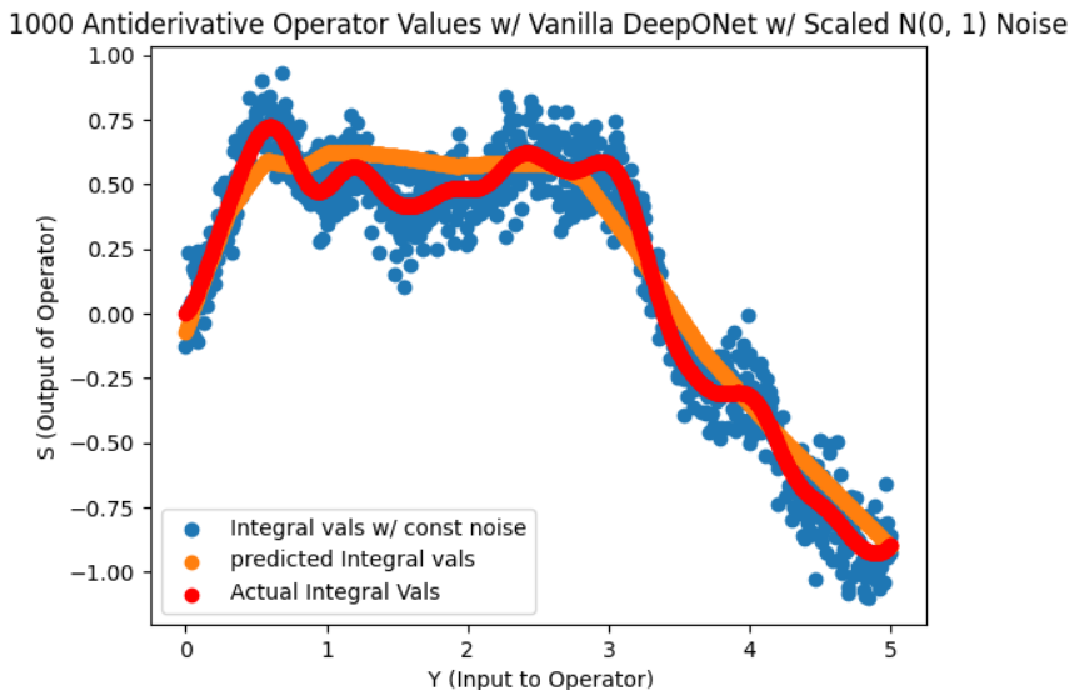}
    \end{minipage}
    \begin{minipage}{0.5\textwidth}
        \includegraphics[scale=0.38]{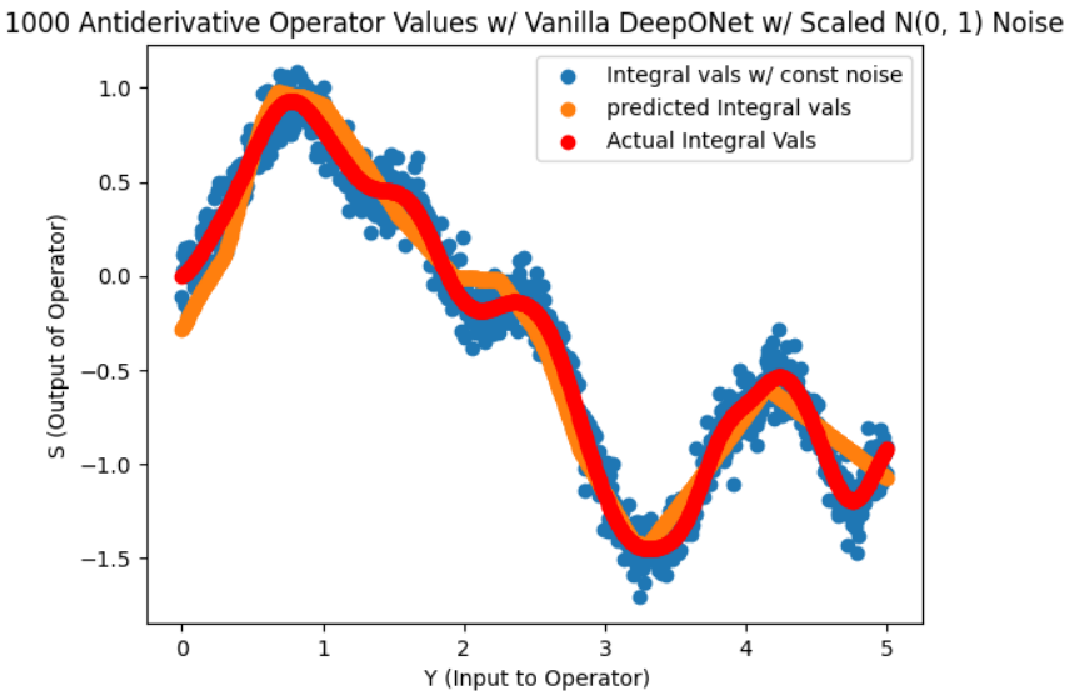}
    \end{minipage}
    \hfill
    \caption{Randomly Selected Antiderivative Operator Output functions with Vanilla DeepONet Predictions.}
    \label{fig:VdeeponetANti}
\end{figure}

For the antiderivative operator with noise (\ref{ex1}), the branch net was a 3 layer sub-network with a consistent 100 hidden neurons per layer while trunk net was a 3 layer sub-network with 64 and 100 hidden neurons per layer. Both had identical neuron output dimensions of 100 each, just like in SON. Vanilla DeepONet achieved final MSEs of 0.016 training and 0.06 testing (Figure \ref{fig:VdeeponetANtiloss}, left), which is comparable to the ending loss for SON. On its own, this would seem to indicate that the vanilla DeepONet is able to effectively learn the noisy antiderivative operator. However, if we examine the operator output functions for randomly selected $u$ as before (Figure \ref{fig:VdeeponetANti}), we see that the regular DeepONet only captures the mean of the noisy operator, and fails to capture the noise level in any way. Performing the scaling factor recover test as before confirms this, with an output noise level of $10^{-7} \approx 0$ for floating point. Additionally, we can see that the vanilla DeepONet fails to capture the finer features of the operator output function that have been obscured by noise. Since the noise scaling factor was independent of the operator output domain $y$, vanilla DeepONet predictions closely align with the actual integral values without noise for many $y$ values. However, it is not clear that this would be the case if the noise scaled differently across the $y$ domain. 

Additionally, vanilla DeepONet's performance was also evaluated for the double integral operator with noise (\ref{DoubleInt}). In this case, the branch net utilized convolutional layers just like applied to the drift and diffusion layer stacks of SON. Vanilla DeepONet achieved final MSEs of 0.002 training and 0.005 testing (Figure \ref{fig:VdeeponetANtiloss}, right), slightly smaller than for SON. However, since SON introduces randomized noise in the forward propagation, MSE loss is restricted by the magnitude of that noise band, depending on what values are randomly sampled by the brownian motion during any given prediction. 

\begin{figure}[h!]
\centering
   \begin{minipage}{\textwidth}  \includegraphics[scale=0.3]{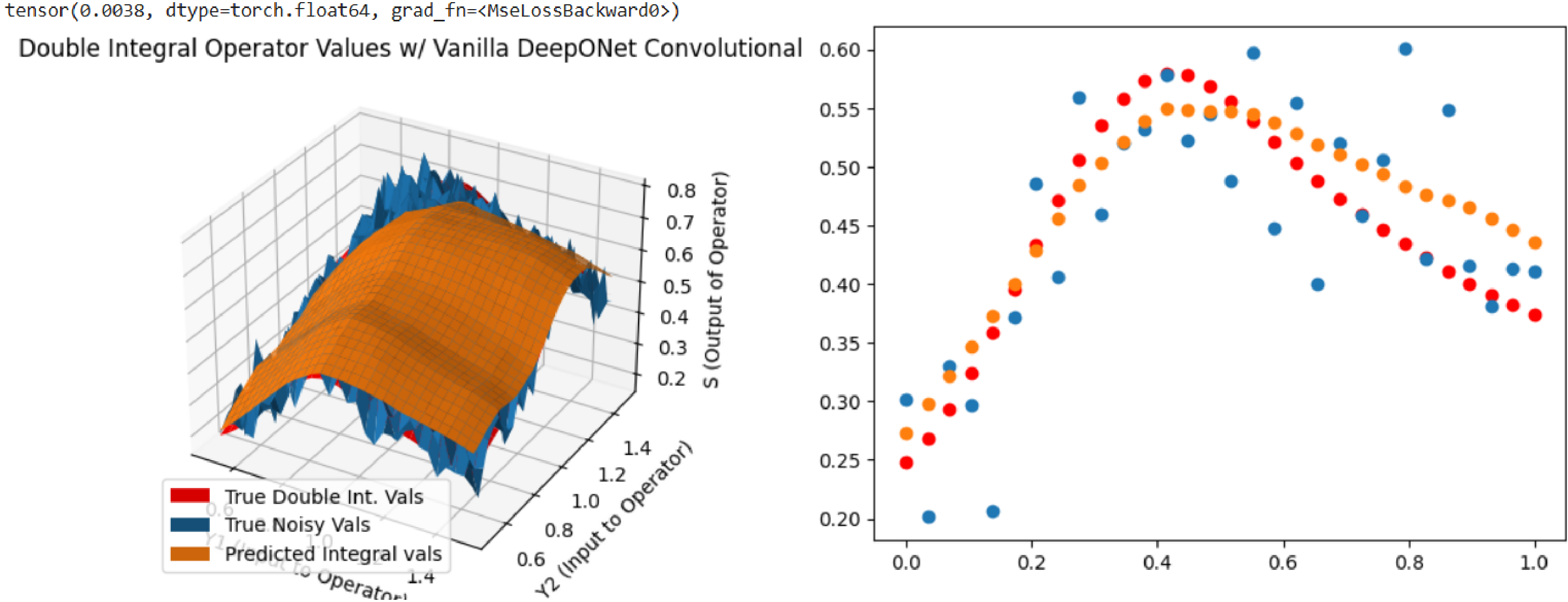}
    \label{fig:pred 1 pair}
\end{minipage}
\begin{minipage}{\textwidth}
 \includegraphics[scale=0.3]{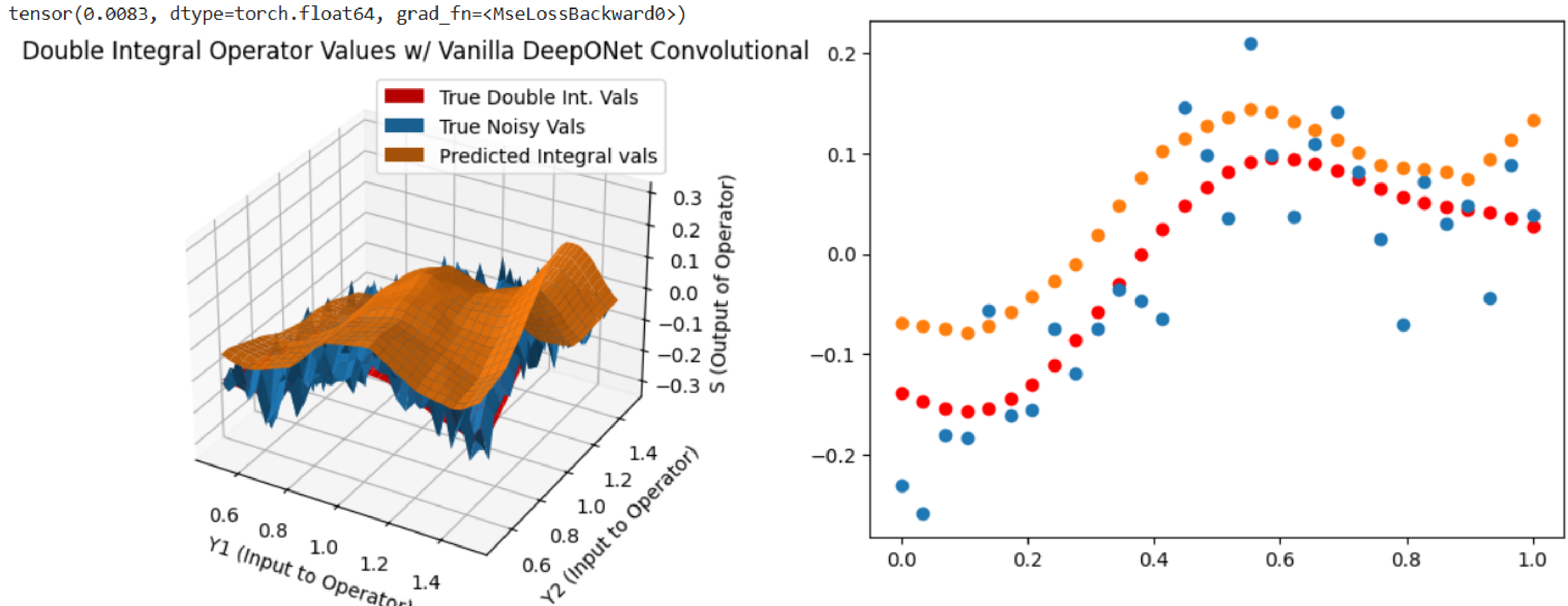}
    \label{fig:pred 2 pair}
\end{minipage}

    \caption{\small Two random Vanilla DeepONet double integral operator with noise test set predictions compared with true operator output (left) and corresponding random cross sections (right).}
    \label{fig:doubleIntegralVDeepONet}
\end{figure}

Additionally, as expected, vanilla DeepONet fails to quantify the noise level of the operator in its predictions (Figure \ref{fig:doubleIntegralVDeepONet}). The DeepONet predictions also struggle to capture the true underlying double integral values, as randomized noise outliers skew the mean of the operator output surface. In Figure \ref{fig:doubleIntegralVDeepONet}, this results in the predicted surfaces being slightly above the true double integral values. While random noise outliers also impacted the SON predictions, the quantification of the noise level in the SNN drift layers allowed SON to produce a correctly scaled noise band that includes the true double integral output across a majority of the surface (Figure \ref{fig:doubleintplots}). 

We also compare the time to train a vanilla DeepONet versus SON across all experiments. Experiments were run on a Google Colab T4 GPU. Training was for 2000 epochs with a full dataset batch size for the antiderivative, ODE, and 2D ODE operators, while training was 100 epochs with a batch size of 900 for the double integral experiments. The networks also had similar numbers of hidden layers and neurons per hidden layer. 

Since SON relies on a much more complicated backpropagation to update the weights, in addition to a more intricate forward process, one might expect it to take significantly longer to train. However, this is not the case, as Table \ref{tab:trainintimecomparison} demonstrates no clear difference between the two methods. Training SON is just as fast as training DeepONet for the same number of epochs. Of course, this does not take into account the performance of each network after a certain number of epochs. SON required 200 epochs to achieve an accurate, convergent performance when predicting the noisy double integral, while the vanilla DeepONet only required 100 to achieve the aforementioned results. Recall however that DeepONet does not quantify noise in the operator while SON does, resulting in a more difficult optimization. 

\begin{table}[]
    \centering
    \begin{tabular}{|c|c|c|c|c|}
    \hline
         & Antiderivative  & ODE & 2D ODE & Double Integral \\
         \hline
        DeepONet & 6:58 & 6:47 &5:29 &13:25 \\
        \hline
         SON & 7:01 & 7:08 &7:26 & 8:05\\
         \hline
    \end{tabular}
    \caption{Training Time Comparison.}
    \label{tab:trainintimecomparison}
\end{table}

\section{Conclusion}

In this work, we presented a novel training mechanism to learn the stochastic functional operators, namely the Stochastic Operator Network (SON). We built on the DeepONet architecture by replacing its branch network with Stochastic Neural Networks (SNNs), which enables explicit quantification of operator uncertainty. The network's training was then formulated as a stochastic optimal control problem, which was solved by utilizing the Stochastic Maximum Principle (SMP). To demonstrate the effectiveness of our approach, we performed numerical experiments on both integral operators and solution operators arising from stochastic differential equations. The results demonstrated that the SON can accurately replicate the outputs of noisy operators in two- and three-dimensional settings while effectively quantifying operator uncertainty via trainable parameters. Moreover, SON consistently recovered the noise-scaling factor across all experiments, illustrating the strength of our approach for uncertainty quantification during training. We also showed that our new method achieved comparable performance to that of the vanilla DeepONet. More specifically, the SON incurs no significant training penalty while accurately replicating operator noise. Future work includes exploring more complex uncertainty settings for functional operators, applying the SON to challenging SDE and SPDE models, and incorporating data assimilation techniques to extend its applicability to scenarios with sparse training observations. 


\bibliographystyle{unsrt}
\bibliography{References}

\begin{thebibliography}{10}

\bibitem{Bhattacharya2021}
K.~Bhattacharya, B.~Hosseini, N.~B. Kovachki, and A.~M. Stuart.
\newblock Model reduction and neural networks for parametric pdes.
\newblock {\em The SMAI Journal of computational mathematics}, 7:121--157,
  2021.

\bibitem{Chen1995}
T.~Chen and H.~Chen.
\newblock Universal approximation to nonlinear operators by neural networks
  with arbitrary activation functions and its application to dynamical systems.
\newblock {\em IEEE transactions on neural netwrosk}, 6(4):911--917, 1995.

\bibitem{Guo2024}
L.~Guo, H.~Wu, Y.~Wang, W.~Zhou, and T.~Zhou.
\newblock Ib-uq: Information bottleneck based uncertainty quantification for
  neural function regression and neural operator learning.
\newblock {\em Journal of Computational Physics}, 510:113089, 2024.

\bibitem{Kovachki2024}
N.~B. Kovachki, S.~Lanthaler, and A.~M. Stuart.
\newblock Operator: Algorithms and analysis.
\newblock arXiv preprint, arXiv:2402.15715, 2024.

\bibitem{Lee2024}
S.~Lee and Y.~Shin.
\newblock On the training and generalization of deep operator networks.
\newblock {\em SIAM Journal on Scientific Computing}, 46(4):C273--C296, 2024.

\bibitem{Li2020}
Z.~Li, N.~Kovachki, K.~Azizzadenesheli, B.~Liu, K.~Bhattacharya, A.~Stuart, and
  A.~Anandkumar.
\newblock Fourier neural operator for parametric partial differential
  equations.
\newblock arXiv preprint, arXiv:2010.08895, 2020.

\bibitem{Lu2022}
L.~Lu, X.~Meng, S.~Cai, Z.~Mao, S.~Goswami, Z.~Zhang, and G.~E. Karniadakis.
\newblock A comprehensive and fair comparison of two neural operators (with
  practical extensions) based on fair data.
\newblock {\em Computer Methods in Applied Mechanics and Engineering},
  393:114778, 2022.

\bibitem{Rahman2022}
Md~A. Rahman, M.~A. Florez, A.~Anandkumar, Z.~E. Ross, and K.~Azizzadenesheli.
\newblock Generative adversarial neural operators.
\newblock arXiv preprint, arXiv:2205.03017, 2022.

\bibitem{Zhang2022}
Z.~Zhang, W.~T. Leung, and H.~Schaeffer.
\newblock Belnet: Basis enhanced learning, a mesh-free neural operator.
\newblock arXiv preprint, arXiv:2212.07336, 2022.

\bibitem{Lu2019}
L.~Lu, P.~Jin, and G.~E. Karniadakis.
\newblock Deeponet: Learning nonlinear operators for identifying differential
  equations based on the universal approximation theorem of operators.
\newblock arXiv preprint, arXiv:1910.03193, 2019.

\bibitem{Lu2021}
L.~Lu, , G.~Pang, P.~Jin, Z.~Zhang, and G.~E. Karniadakis.
\newblock Learning nonlinear operators via deeponet based on the universal
  approximation theorem of operators.
\newblock {\em Nat. Mach. Intell.}, 3:218--229, 2021.

\bibitem{Chen2018}
R.T.~Q. Chen, Y.~Rubanova, J.~Bettencourt, and D.~Duvenaud.
\newblock Neural ordinary differential equations.
\newblock In {\em Proceedings of the 32nd International Conference on Neural
  Information Processing Systems}, NIPS'18, page 6572–6583, Red Hook, NY,
  USA, 2018. Curran Associates Inc.

\bibitem{Dupont2019}
E.~Dupont, A.~Doucet, and Y.~W. Teh.
\newblock Augmented neural odes.
\newblock In {\em Advances in Neural Information Processing Systems},
  volume~32, pages 3140--3150, Red Hook, NY, 2019. Curran Associates.

\bibitem{Gerstberger1997}
R.~Gerstberger and P.~Rentrop.
\newblock Feedforward neural nets as discretization schemes for odes and daes.
\newblock {\em J. Comput. Appl. Math.}, 82:117--128{}, 1997.

\bibitem{Haber2018}
E.~Haber and L.~Ruthotto.
\newblock Stable architectures for deep neural networks.
\newblock {\em Inverse Problems}, 34:014004, 2018.

\bibitem{Weinan2017}
E.~Weinan.
\newblock A propsal on machine learning via dynamical systems.
\newblock {\em Commun. Math. Stat.}, 5:1--11, 2017.

\bibitem{Jia2019}
J.~Jia and A.~Benson.
\newblock Neural jump stochastic differential euquations.
\newblock In {\em 33rd Conference on Neural Information Processing Systems},
  2019.

\bibitem{Kong2020}
L.~Kong, J.~Sun, and C.~Zhang.
\newblock Sde-net: Equipping deep neural networks with uncertainty estimates.
\newblock In {\em Proceedings of the 37th International Conference on Machine
  Learning}, 2020.

\bibitem{Liu2019}
X.~Liu, T.~Xiao, S.~Si, Q.~Cao andS. K.~Kumar, and C.-J. Hsieh.
\newblock Neural sde: Stabilizing neural ode networks with stochastic noise.
\newblock arXiv preprint, 2019.

\bibitem{Liu2020}
X.~Liu, T.~Xiao, S.~Si, Q.~Cao, S.~Kumar, and C.-J. Hsieh.
\newblock How does noise help robustness? explanation and exploration under the
  neural sde framework.
\newblock In {\em Proceedings of the IEEE/CVF Conference on Computer Vision and
  Pattern Recognition (CVPR)}, 2020.

\bibitem{Tzen2019}
B.~Tzen and M.~Raginsky.
\newblock Neural stochastic differential equations: Deep latent gaussian models
  in the diffusion limit.
\newblock arXiv: Learning.

\bibitem{Geneva2019}
N.~Geneva and N.~Zabaras.
\newblock Quantifying model form uncertainty in reynolds-averaged turbulence
  models with bayesian deep neural networks.
\newblock {\em J. Comput. Phys.}, 383:125--147, 2019.

\bibitem{Geneva2020}
N.~Geneva and N.~Zabaras.
\newblock Modeling the dynamics of pde systems with physics-constrained deep
  auto-regressive networks.
\newblock {\em J. Comput. Phys.}, 403:109056, 2020.

\bibitem{Kwon2020}
Y.~Kwon, J.-H. Won, B.~J. Kim, and M.~C. Pail.
\newblock Uncertainty quantification using bayesian neural networks in
  classification: Application to biomedical image segmentation.
\newblock {\em Comput. Statist. Data Anal.}, 142:106816, 2020.

\bibitem{McDermott2019}
P.~L. McDermott and C.~K. Wikle.
\newblock Bayesian recurrent neural network models for forecasting and
  quantifying uncertainty in spatial-temporal data.
\newblock {\em Entropy}, 21(2):184, 2019.

\bibitem{Savchenko2020}
A.~V. Savchenko.
\newblock Probabilistic neural network with complex exponential activation
  functions in image recognition.
\newblock {\em IEEE Trans. Neural Netw. Learn. Syst.}, 31:651--660, 2020.

\bibitem{Wu2020}
L.~Wu, K.~Zulueta, Z.~Major, A.~Arriaga, and L.~Noels.
\newblock Bayesian inference of non-linear multiscale model parameters
  accelerated by a deep neural network.
\newblock {\em Comput. Methods Appl. Mech. Engrg.}, 360:112693, 2020.

\bibitem{Yang2021}
L.~Yang, X.~Meng, and G.~E. Karniadakis.
\newblock B-pinns: Bayesian physics-informed neural networks for forward and
  inverse pde problems with noisy data.
\newblock {\em J. Comput. Phys.}, 425:109913, 2021.

\bibitem{Yao2019}
J.~Yao, W.~Pan, S.~Ghosh, and F.~Doshi-Velez.
\newblock Uncertainty quantification for bayesian neural network inference.
\newblock In {\em Proceedings at the International Conference on Machine
  Learning: Workshop on Uncertainty \& Robustness in Deep Learning (ICML)},
  2019.

\bibitem{Archibald2024}
R.~Archibald, F.~Bao, Y.~Cao, and H.~Sun.
\newblock Numerical analysis for convergence of a sample-wise backpropagation
  method for training stochastic neural networks.
\newblock {\em SIAM Journal on Numerical Analysis}, 62(2):593--621, 2024.

\bibitem{Bao2022}
R.~Archibald, F.~Bao, Y.~Cao, and H.~Zhang.
\newblock A backward sde method for uncertainty quantification in deep
  learning.
\newblock {\em Discrete and Continuous Dynamical Systems - S},
  15(10):2807--2835, 2022.

\bibitem{Bao2020a}
R.~Archibald, F.~Bao, and J.~Yong.
\newblock A stochastic gradient descent approach for stochastic optimal
  control.
\newblock {\em East Asian Journal on Applied Mathematics}, 10(4):635--658,
  2020.

\bibitem{Andersson2009}
D.~Andersson.
\newblock {\em Contributions to the Stochastic Maximum Principle}.
\newblock PhD thesis, KTH Royal Institute of Technology, Sweden, 2009.

\bibitem{Ma1999}
J.~Ma and J.~Yong.
\newblock {\em Forward-backward stochastic differential equations and their
  applications}.
\newblock Number 1702. Springer Science $\&$ Business Media, 1999.

\bibitem{Bottou2018}
L.~Bottou, F.~E. Curtis, and J.~Nocedal.
\newblock Optimization methods for large-scale machine learning.
\newblock {\em SIAM Review}, 60(2):223--311, 2018.

\bibitem{bao2016first}
Feng Bao, Yanzhao Cao, Amnon Meir, and Weidong Zhao.
\newblock A first order scheme for backward doubly stochastic differential
  equations.
\newblock {\em SIAM/ASA Journal on Uncertainty Quantification}, 4(1):413--445,
  2016.

\bibitem{Bao_DA_BSDE}
Feng Bao, Yanzhao Cao, and Jiongmin Yong.
\newblock Data informed solution estimation for forward-backward stochastic
  differential equations.
\newblock {\em Analysis and Applications}, 19(3):439--464, 2021.

\bibitem{Bjork2019}
T.~Bj\"{o}rk.
\newblock {\em Arbitrage Theory in Continuous Time}.
\newblock Oxford University Press, 12 2019.

\bibitem{Bao2018}
F.~Bao, Y.~Cao, and W.~Zhao.
\newblock A backward doubly stochastic differential equation approach for
  nonlinear filtering problems.
\newblock {\em Commun. Comput. Phys.}, 23(5):1573--1601, 2018.

\bibitem{VAnh2024}
V.-A. Le and M.~Dik.
\newblock A mathematical analysis of neural operator behaviors.
\newblock arXiv preprint arXiv:2410.21481, 2024.

\end{thebibliography}

\end{document}